\documentclass[accepted]{uai2024} 
                        
\usepackage[american]{babel}

\usepackage{natbib} 
\usepackage{mathtools} 
\usepackage{booktabs} 
\usepackage{tikz} 

\usepackage{microtype}
\usepackage{graphicx}
\usepackage{subcaption}

\usepackage{comment}

\usepackage{amsmath}
\usepackage{amssymb}
\usepackage{mathtools}
\usepackage{amsthm}
\usepackage{scalerel}

\usepackage[capitalize,noabbrev]{cleveref}

\makeatletter
\newtheorem*{rep@theorem}{\rep@title}
\newcommand{\newreptheorem}[2]{%
\newenvironment{rep#1}[1]{%
 \def\rep@title{#2 \ref{##1}}%
 \begin{rep@theorem}}%
 {\end{rep@theorem}}}
\makeatother

\newcounter{set}
\newreptheorem{proposition}{Proposition}

\newreptheorem{corollary}{Corollary}
\newreptheorem{theorem}{Theorem}
\newreptheorem{lemma}{Lemma}
\newreptheorem{observation}{Observation}
\newreptheorem{remark}{Remark}

\newtheorem{observation}{Observation}

\theoremstyle{plain}
\newtheorem{theorem}{Theorem}[section]
\newtheorem{proposition}[theorem]{Proposition}
\newtheorem{lemma}[theorem]{Lemma}
\newtheorem{corollary}[theorem]{Corollary}
\theoremstyle{definition}
\newtheorem{definition}[theorem]{Definition}
\newtheorem{assumption}[theorem]{Assumption}
\theoremstyle{remark}
\newtheorem{remark}[theorem]{Remark}

\usepackage{amsmath,amsfonts,bm}
\usepackage{mathrsfs}

\def\eqref#1{equation~\ref{#1}}

\def\1{\bm{1}}

\def\eps{{\epsilon}}

\def\law{{\mathrm{Law}}}

\def\ve{{\bm{e}}}

\DeclareMathAlphabet{\mathsfit}{\encodingdefault}{\sfdefault}{m}{sl}
\SetMathAlphabet{\mathsfit}{bold}{\encodingdefault}{\sfdefault}{bx}{n}

\def\gA{{\mathcal{A}}}

\def\gG{{\mathcal{G}}}

\def\gN{{\mathcal{N}}}
\def\gO{{\mathcal{O}}}

\def\sR{{\mathbb{R}}}

\newcommand{\E}{\mathbb{E}}

\newcommand{\R}{\mathbb{R}}

\let\log\relax
\DeclareMathOperator{\log}{ln}

\def\R{{\mathbb{R}}}

\def\dd{{\mathrm{d}}}

\newcommand{\KL}{\mathrm{KL}}

\DeclareMathOperator*{\argmin}{arg\,min}

\title{To smooth a cloud or to pin it down: Guarantees and Insights on Score Matching in Denoising Diffusion Models}

\author[1]{\href{mailto:<teodroa.reu@balliol.ox.ac.uk>?Subject=Your UAI 2024 paper}{Teodora Reu \textsuperscript{\ensuremath{*}} }{}}
\author[2]{Francisco Vargas \textsuperscript{\ensuremath{*}} }
\author[2]{Anna Kerekes}
\author[1]{Michael M. Bronstein}
\affil[1]{%
    University of Oxford\\
}
\affil[2]{%
    University of Cambridge\\
}

\begin{document}
\maketitle

\begin{NoHyper}
\renewcommand{\thefootnote}{\ensuremath{*}}
\footnotetext{Equal contribution.}
\end{NoHyper}

\begin{abstract}
Denoising diffusion models are a class of generative models that have recently achieved state-of-the-art results across many domains. Gradual noise is added to the data using a diffusion process, which transforms the data distribution into a Gaussian. Samples from the generative model are then obtained by simulating an approximation of the time reversal of this diffusion initialized by Gaussian samples. Recent research has explored the sampling error achieved by diffusion models under the assumption of an absolute error $\epsilon$ achieved via a neural approximation of the score. To the best of our knowledge, no work formally quantifies the error of such neural approximation to the score. In this paper, we close the gap and present quantitative error bounds for approximating the score of denoising diffusion models using neural networks leveraging ideas from stochastic control. Finally, through simulation, we explore some of the insights that arise from our results confirming that diffusion models based on the Ornstein-Uhlenbeck (OU) process require fewer parameters to better approximate the score than those based on the F\"{o}lmer drift / Pinned Brownian Motion.
\end{abstract}

\section{Introduction}
\label{sec:intro}

Let $\pi$ be a probability density on $\mathbb{R}^d$ of the form
\begin{equation}
    \pi(x)=\frac{\gamma(x)}{Z},\qquad Z=\int_{\mathbb{R}^d} \gamma(x) \mathrm{d}x,
\end{equation}
where $\gamma:\mathbb{R}^d \rightarrow \mathbb{R}^{+}$ can be evaluated pointwise but the normalizing constant $Z$ is intractable. In both the sampling problem and generative modelling, one is interested in obtaining approximate samples from $\pi$.  In sampling, one has access to $\gamma$ whilst in generative modelling we only have access to samples from $x_i \sim  \pi(x)$. 

While superficially similar, methodologies for these two different tasks initially evolved quite separately. Due to the ability to take gradient sampling, a variety of Markov Chain Monte Carlo (MCMC) \citep{neal2011mcmc}, as well as variational \citep{wainwright2008graphical, blei2017variational} techniques have been developed to tackle the sampling problem. In variational techniques, one considers a flexible family of easy-to-sample distributions $q^{\theta}$ whose parameters are optimized by minimizing a suitable cost, such as reverse Kullback--Leibler discrepancy $\KL(q^{\theta}||\pi)$.



Complementary, generative modelling is interested in being able to sample from the underlying density $\pi$ when only a set of finite samples is available. As a result, most methodologies were initially based on forward KL (i.e. Maximum Likelihood) like approaches, where one trains a tractable model $q^{\theta}$ via minimizing $\KL(\pi || q^{\theta})$ \citep{papamakarios2019normalizing} which can be achieved as we can estimate gradients $\nabla_\theta\KL(\pi || q^{\theta})$ using samples from $\pi$.

Recent score-based techniques for generative modelling \citep{song2020score} constitute of nice cross-pollination between the standard techniques used in sampling (e.g. MCMC) ported over to generative modelling and in some cases feeding back into the sampling community \citep{doucet2022annealed,vargas2023denoising}. 

In recent years we have seen the rise of Denoising Diffusion Probabilistic Models (DDPM), a powerful class of generative models \citep{sohl2015deep,ho2020denoising,song2020score} to sample from unnormalized densities. In this context, one adds noise progressively to data using diffusion to transform the complex target distribution into a Gaussian distribution. The time reversal of this diffusion can then be used to transform a Gaussian sample into a sample from the target. As with many theoretical works pertaining to diffusion models \citep{chen2022sampling, lee2023convergence}, we will assume the target distribution admits a density for our analysis; this is a common assumption in the analysis of sampling algorithms \citep{ma2019sampling,vempala2019rapid} and is not restrictive.

It is important to highlight that diffusion models have also recently made it into sampling \citep{vargas2023denoising,berner2022optimal}, in particular, these works establish connections between DDPM and well-established field of stochastic control \citep{kappen2012optimal,nusken2021solving}.


In this work, we delve into the connection to stochastic control highlighted among denoising diffusion models \citep{ho2020denoising,song2020score} in \cite{vargas2023denoising}. We leverage this connection to show how the score of VP-SDEs \citep{song2020score} can be approximated with neural networks up to an arbitrarily small error, and we quantify the induced sampling error.

Our contributions in this paper can be summarized as follows: 

\begin{itemize}
    \item Establishing a connection between the VP-SDE score and OU-semigroup (Section \ref{ousec}). 
    \item  Exploring novel regularity properties for OU-semigroup (Section \ref{sec:regularity}), via leveraging connections to stochastic control \citep{tzen2019theoretical}.
    \item Demonstrating neural network and sampling approximation results for a simplified VP-SDE (Proposition \ref{col:est}, Remark \ref{rem:approx}) with minimal assumptions on the data/target distribution $\pi$.
    \item Leveraging some of the insights/conjectures motivated by our theoretical results we carry out a set of empirical explorations contrasting two different types of approaches of SDEs for score-based generative modelling (F\"ollmer Drift vs VP-SDE).
\end{itemize}

Upon comparing our approach with related works, notable differences in assumptions emerge. For instance,  \cite{chen2023score} make manifold assumptions regarding data representation, diverging substantially from our less restrictive assumptions. Additionally, their methodology employs covers based on score networks rather than the OU-semigroup, which contrasts with ours. Similarly, in \cite{oko2023diffusion,cole2024score}, assumptions about data distribution and analysis methods differ notably from ours, with their analysis focusing on score loss rather than the OU-semigroup. We highlight that whilst the concurrent work in \cite{cole2024score} has similar assumptions on the target distribution, their focus is strictly on the data-driven case, and whilst some overlap in the spirit of the sketch, they obtain different bounds. 

\section{Background - Denoising Diffusion Models and Stochastic Control}\label{sec:DDSCT}
For this work, we will introduce Denoising Diffusions in continuous time. Let $\mathcal{C}=C([0,T],\mathbb{R}^d)$ be the space of continuous functions from $[0,T]$ to $\mathbb{R}^d$ and $\mathcal{B}(\mathcal{C})$ the Borel sets on $\mathcal{C}$. We consider path measures, which are probability measures on $(\mathcal{C},\mathcal{B}(\mathcal{C}))$ \citep{leonard2013survey}. To relate denoising diffusion models to the methodology in \cite{tzen2019theoretical}, we will introduce connections presented in \cite{vargas2023denoising}, which relate score matching in VP-SDEs to stochastic control, thus enabling our main results.



\begin{figure*}[t!]
    \centering 
    \begin{subfigure}[t]{0.49\textwidth}
        \centering
        \includegraphics[width=\textwidth]{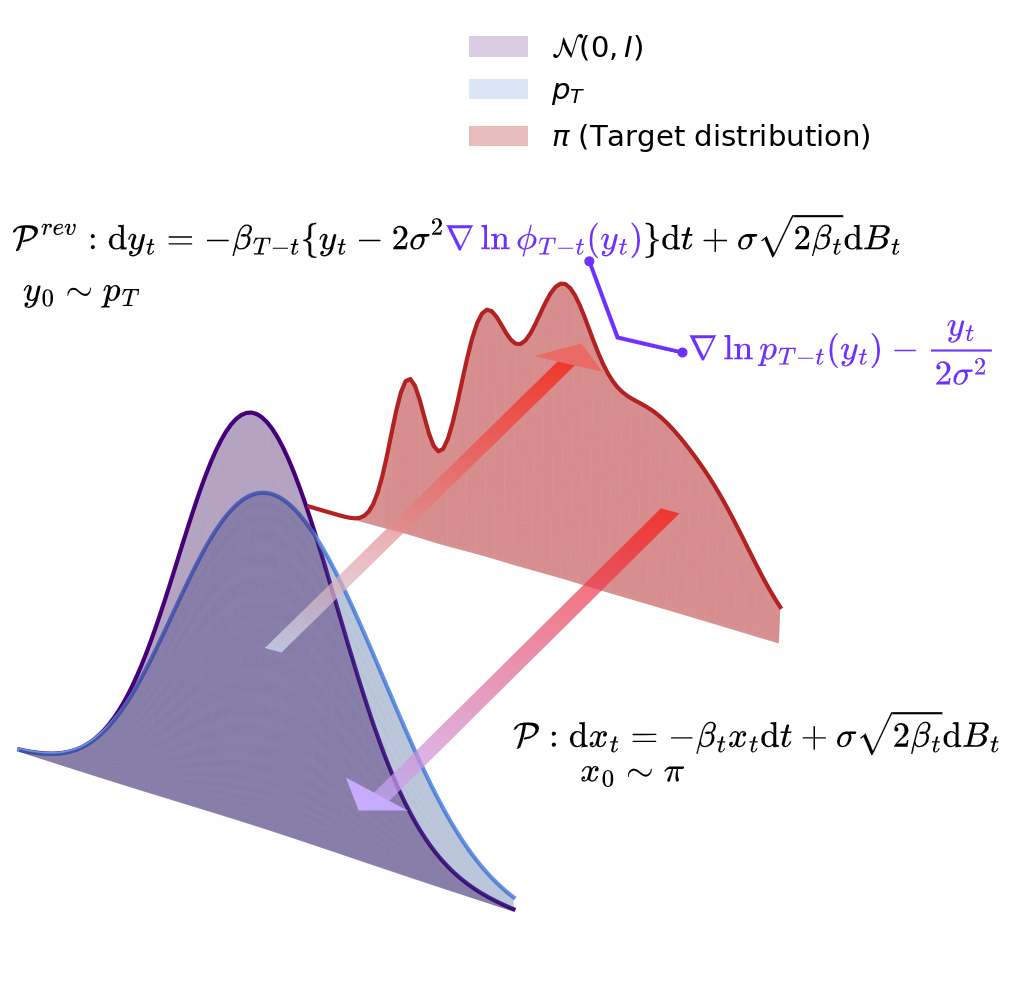}
        \caption{}
    \end{subfigure}%
    \begin{subfigure}[t]{0.49\textwidth}
        \centering
       \includegraphics[width=\textwidth]{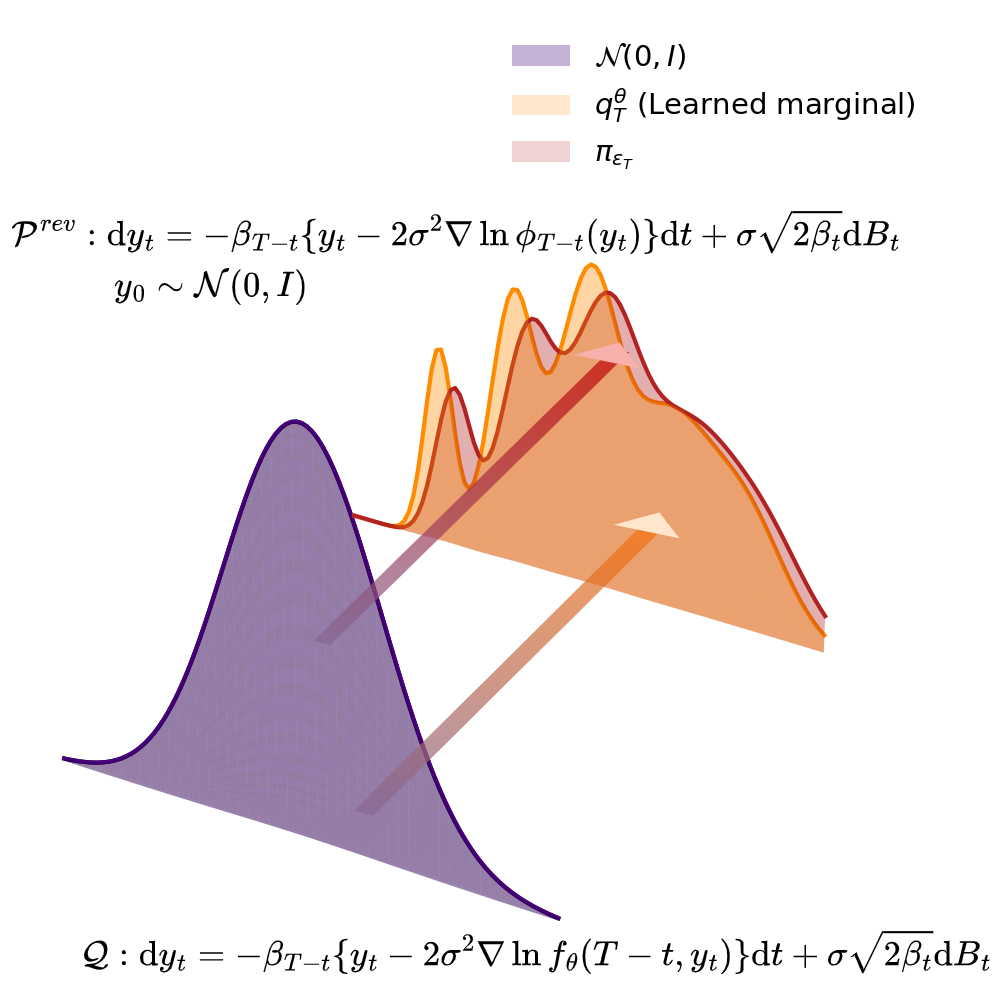}
        \caption{}
    \end{subfigure}
    \caption{ a) Noise-adding process for exact reversal. The distribution  $\gN(0,I)$ is drawn for comparison to $p_T$.  b)Exact and approximate time reversal starting from $\gN(0,I)$ the former exhibits only the mixing error whilst the latter incorporates the network's approximation error.} \label{fig:pic}
\end{figure*}

\subsection{Backwards diffusion and its time-reversal}
Consider the forward noising diffusion given by a time-reversed Ornstein--Uhlenbeck (OU) process (\cite{song2020score} refer to this SDE as the VP-SDE):
\begin{equation}\label{eq:forwarddiffusionP}
    \mathrm{d}x_t=-\beta_t x_t \mathrm{d}t+\sigma \sqrt{2\beta_t}\mathrm{d}B_t,\qquad x_0 \sim \pi, 
\end{equation}
where $(B_t)_{t\in[0,T]}$ is a $d$-dimension Brownian motion and $t \rightarrow \beta_t$ is a non-decreasing positive function. This diffusion induces the path-measure $\mathcal{P}$ on the time interval $[0,T]$, and the marginal density of $x_t$ is denoted $p_t$. The transition density of this diffusion is given by $p_{t|0}(x_t|x_0)=\mathcal{N}(x_t;\sqrt{1-\lambda_t}x_0,\sigma^2 \lambda_t I)$, where $\lambda_t=1-\exp(-2\int^t_0\beta_s \mathrm{d}s)$. We will always consider a scenario where  $\int_0^T \beta_s \mathrm{d}s \gg 1$ so that $p_T(x)\approx \mathcal{N}(x;0,\sigma^2 I)$. 

From \citep{haussmann1986time},  its time-reversal $(y_t)_{t\in[0,T]}=(x_{T-t})_{t\in[0,T]}$, where equality is here in distribution, yields the forward time diffusion:
\begin{align}
\label{eq:exacttimereversalCT}
{\mathrm{d}y_t} &{= \beta_{T-t}\{y_t+2\sigma^2 \nabla \log p_{T-t}(y_t)\} \mathrm{d}t \nonumber }\\
&{\quad + \sigma \sqrt{2\beta_{T-t}}\mathrm{d}W_t, \quad y_0 \sim p_{T},}
\end{align}
where $(W_t)_{t\in[0,T]}$ is another $d$-dimensional Brownian motion. By definition this time-reversal starts from $y_0 \sim p_T = \law(x_t) \approx \mathcal{N}(0,\sigma^2 I)$ and is such that $y_T \sim \pi$. This suggests that approximate simulation of diffusion (\ref{eq:exacttimereversalCT}) would result in approximate samples from $\pi$.  However, putting this idea into practice requires being able to approximate the intractable scores $(\nabla \log p_t(x))_{t \in [0,T]}$. Unlike DDPM, score matching techniques are not feasible, as sampling from (\ref{eq:forwarddiffusionP}) requires sampling $x_0 \sim \pi$, which is impossible by assumption.

\subsection{Reference diffusion and value function}\label{sec:refdiffusionvaluefunction}
To introduce the value function, it is first useful to introduce a \emph{reference} process defined by the diffusion following (\ref{eq:forwarddiffusionP}), but initialized at {$p^{\textup{ref}}_0(x)=\mathcal{N}(x;0,\sigma^2 I)$} rather than $\pi(x)$ thus ensuring that the marginals of the resulting path measure $\mathcal{P}^{\textup{ref}}$ all satisfy $p^{\textup{ref}}_t(x)=\mathcal{N}(x;0,\sigma^2 I)$. Following \cite{vargas2023denoising} we can identify $\mathcal{P}$  as the path measure minimizing the half bridge $\mathcal{P}= \argmin_\mathcal{Q} \{\KL(\mathcal{Q}||\mathcal{P}^{\textup{ref}}): q_T=\pi\}$ \citep{bernton2019SBsamplers,vargasshro2021, debortoli2021diffusion}.
where representation of $\mathcal{P}^{\textup{ref}}$ is given by 
\begin{equation}\label{eq:timereversalrefprocessCT}
    \mathrm{d}y_t =-\beta_{T-t} y_t \mathrm{d}t+\sigma \sqrt{2\beta_{T-t}} \mathrm{d}W_t,\qquad y_0 \sim p^{\textup{ref}}_0.
\end{equation}
Noting that $\beta_{T-t} (y_t +2\sigma^2 \nabla \log p^{\textup{ref}}_{t}(y_t))=-\beta_{T-t} y_t$, \cite{vargas2023denoising} rewrites the time-reversal (\ref{eq:exacttimereversalCT}) of $\mathcal{P}$ as 
\begin{align}
\label{eq:diffusionvaluefunction}
\mathrm{d}y_t &= -\beta_{T-t} \{y_t -2\sigma^2 \nabla \log \phi_{T-t}(y_t)\} \mathrm{d}t \nonumber \\
&\quad + \sigma \sqrt{2\beta_{T-t}} \mathrm{d}W_t, \quad y_0 \sim p_T,
\end{align}
where $v_t(x) = -\ln \phi_t(x)=-\ln p_t(x)/p^{\textup{ref}}_t(x)$ is known as the value function \citep{fleming2012deterministic,Pham:2009,nusken2021solving,tzen2019theoretical}. We point the reader to Figure \ref{fig:pic} for a pictorial illustration of the aforementioned reversal and value function.

\subsection{Learning the Forward Diffusion - Reverse KL / Stochastic Control Formulation}
To approximate  (\ref{eq:exacttimereversalCT}), consider a path measure $\mathcal{Q}^\theta$ which is induced by
\begin{align}
\label{eq:Qthetascore}
\mathrm{d}y_t &= \beta_{T-t}\{y_t+2\sigma^2 s_{\theta}(t,y_t) \} \mathrm{d}t \nonumber \\
&\quad + \sigma \sqrt{2\beta_{T-t}}\mathrm{d}W_t, \quad y_0  \sim \mathcal{N}(0,\sigma^2 I),
\end{align}
so that $y_t \sim q^{\theta}_{t}$. To obtain $s_{\theta}(t,x) \approx \nabla \log p_t(x)$, we parameterize $s_{\theta}(t,x)$ by a neural network whose parameters are obtained by minimizing
\begin{multline}
\KL(\mathcal{Q}^\theta||\mathcal{P}) = \KL(\mathcal{N}(0,\sigma^2 I)||p_T) \\
+ \sigma^2 \mathbb{E}_{\mathcal{Q}^\theta}\Biggl[\int_0^T \beta_{T-t}||s_\theta(T-t,y_t)- \nabla \log p_{T-t}(y_t)||^2 \mathrm{d}t \Biggr]. \nonumber
\end{multline}
This expression closely resembles the expression obtained in Theorem 1 of \cite{song2021maximum}. Note, in the DDPM \cite{ho2020denoising} setting as we have samples from $\mathcal{P}$ via simulating the forward SDE then one can recover the score-matching objective from \cite{song2020score},
\begin{multline}
\KL(\mathcal{P} || \mathcal{Q}^\theta) = \KL(\mathcal{N}(0,\sigma^2 I)||p_T)\\
+ \sigma^2 \mathbb{E}_{\mathcal{P}}\Biggl[\int_0^T \beta_{T-t}||s_\theta(T-t,y_t)- \nabla \log p_{T-t}(y_t)||^2 \mathrm{d}t \Biggr]. \nonumber
\end{multline}

To fully make the connection to stochastic control \citep{tzen2019theoretical,dai1991stochastic}, following \cite{vargas2023denoising} we equivalently reparameterize $\mathcal{Q}^\theta$ via the value function formulation of the backward SDE (Equation \ref{eq:diffusionvaluefunction}),
\begin{multline}\label{eq:approximatetimereversalCT}
    \mathrm{d}y_t=-\beta_{T-t}\{y_t -2\sigma^2 f_{\theta}(T-t,y_t)\} \mathrm{d}t \\
    +\sigma \sqrt{2\beta_{T-t}} \mathrm{d}W_t,\qquad y_0 \sim \mathcal{N}(0,\sigma^2 I),
\end{multline}
unlike Equation \ref{eq:Qthetascore} $f_\theta$ approximates $\nabla \ln \phi_t$ rather than the score $\nabla \ln p_t$. Then under this reparameterization \cite{vargas2023denoising} use standard results on half bridges \citep{bernton2019SBsamplers} we can re-express  $\KL(\mathcal{Q}^\theta||\mathcal{P})$ in the following form:
\begin{align}
\KL(\mathcal{Q}^\theta||\mathcal{P})&
=\mathbb{E}_{\mathcal{Q}^\theta} \Biggl[ \sigma^2 \scaleobj{.8}{\int_0^T} \beta_{T-t} ||f_\theta(T-t,y_t)||^2 \mathrm{d}t \nonumber \\
&\quad +\scaleobj{.8}{\ln \left(\frac{\gN(y_T; 0, \sigma^2 I)}{\pi(y_T)}\right)} \Biggr]\label{eq:KLpathintegral},
\end{align}
where $q^{\theta^*}_0 = p_T \approx \gN(0,\sigma^2I)$ \footnote{ $q^{\theta^*}_0$ denotes the optimal  distribution at $t=0$ minimising (\ref{eq:KLpathintegral})}. Then $\theta$ minimizing (\ref{eq:KLpathintegral}), approximate samples from $\pi$ can be obtained by simulating (\ref{eq:approximatetimereversalCT}) and returning $y_T\! \sim \!q^{\theta}_T$.  Note concurrent work \citep{berner2022optimal} also optimizes an equivalent reverse KL to Equation \ref{eq:KLpathintegral}. Equation \ref{eq:KLpathintegral} is an instance of stochastic control \citep{kappen2012optimal,tzen2019theoretical,nusken2021solving,berner2022optimal} akin to the objective studied in \citep{tzen2019theoretical}, these re-formulations as a stochastic control problem, in particular, the connection to the value function (see Remark \ref{rem:ou}) will allow us to provide expresiveness remarks for VP-SDE based diffusions.

\subsection{Pinned Brownian Motion Generative Models and Samplers}\label{sec:pbmgm}

In this section, we reintroduce the class of generative models and samplers studied in \citep{tzen2019theoretical} and highlight the similarities and differences in contrast to the OU-based diffusion models.

The pinned Brownian motion SDE is arrived at by using the h-transform to condition the scaled Brownian motion
\begin{equation}\label{eq:scaleb}
    \mathrm{d}x_t=\sqrt{\frac{\mathrm{d}\alpha_t}{\mathrm{d}t}}\mathrm{d}B_t,\qquad x_0 \sim \pi, 
\end{equation}
 to hit the value $0$ at time $T$, resulting in the forward SDE:
\begin{equation}\label{eq:forwardpbm}
    \mathrm{d}x_t=-\frac{\mathrm{d}\alpha_t}{\mathrm{d}t}\frac{x_t}{\alpha_T -\alpha_t} \mathrm{d}t+ \sqrt{\frac{\mathrm{d}\alpha_t}{\mathrm{d}t}}\mathrm{d}B_t,\quad x_0 \sim \pi, 
\end{equation}
where $x_{T}=0$, furthermore the SDE in Equation \ref{eq:forwardpbm} has the following transition density (full derivation in Appendix \ref{apdx:PBM}):
\begin{equation}
p(x_t | x_0) = \mathcal{N}\left(x_t\Bigg| \frac{\alpha_T-\alpha_t }{\alpha_T-\alpha_0} x_0  , \frac{(\alpha_T-\alpha_t) (\alpha_t - \alpha_0)}{\alpha_T-\alpha_0}  \right), \nonumber
\end{equation}
which we can use to learn the score \citep{song2020score}. Once we have the score the time reversal of Equation \ref{eq:forwardpbm}, yields an SDE which we can use for generative modelling
\begin{align}
\label{eq:exacttimereversalPBM}
\mathrm{d}y_t &= \frac{\mathrm{d}\alpha_{T-t}}{\mathrm{d}t}\Big\{\frac{y_t}{\alpha_T - \alpha_{T-t}}+ \nabla \log p_{T-t}(y_t)\Big\} \mathrm{d}t \nonumber \\
&\quad + \sqrt{\frac{\mathrm{d}\alpha_{T-t}}{\mathrm{d}t}}\mathrm{d}W_t, \quad y_0 =0.
\end{align}
We will refer to this SDE as the backward pinned brownian motion (BPBM). As we will discuss the BPBM SDE is a Generalisation of the F\"ollmer process \citep{dai1991stochastic}, which is a well-studied SDE in stochastic control \citep{dai1991stochastic,kappen2012optimal,tzen2019theoretical,many-paths,vargas2021bayesian}.

Prior work, such as aligned Schrodinger bridges \citep{somnath2023aligned,liu20232}, have discussed this SDE in the context of dataset alignment and conditional generative modelling. Additionally, First Hitting Diffusion models \citep{ye2022first} have explored a variant of PBM where instead of a Brownian motion Equation \ref{eq:scaleb} is replaced with a VP-SDE. However, to our knowledge, PBM has not yet been compared carefully to VP-SDE within the context of generative modelling (some comparison has been done empirically for sampling \citep{vargas2023denoising,berner2022optimal}).

\section{Expressiveness and Regularity Results}

In this section, we present our main result. We demonstrate that $\nabla \ln \phi_t$ and thus the score of the OU-SDE can be approximated by a multi-layer neural network efficiently.

Theorem 3.1 in \cite{tzen2019theoretical} provides neural network approximation and sampling guarantees for a different class of SDEs than DDPM (i.e. Equations \ref{eq:exacttimereversalCT} or \ref{eq:diffusionvaluefunction}). Thus in this section, we will adapt such results to denoising diffusion samplers \citep{vargas2023denoising} and via directly relating the approximations to the score of the VP-SDE (Equation \ref{eq:forwarddiffusionP}) we motivate how these results extend to DDPM based methods \citep{song2020score, ho2020denoising,huang2021variational}.

\cite{tzen2019theoretical} guarantee approximate sampling from a target distribution using a multilayer feedforward neural net drift, assuming the smoothness, Lipschitzness, and boundedness of $f(x)=\frac{\mathrm{d}\pi}{\mathrm{d}\gN(0, \sigma^2 I)}(x)$, (Assumption \ref{assump:a1}), as well as the smoothness of the activations (Assumption \ref{assump:a2}) and uniform approximability of $f$ and its gradient by a neural network (Assumption \ref{assump:a3}). In the following proposition and remark, we present our adaption of their results to DDS.

\begin{proposition} \label{col:est}
Suppose Assumptions in Appendix \ref{assump} are in force. Let L denote the maximum of the Lipschitz constants of $f$ and $\nabla f$. Then for all $0< \epsilon < 16L^2/c^2$, there exists a neural net $\hat{v} : \mathbb{R}^d \times [0,1] \to \mathbb{R}^d$ with size polynomial in $1/\epsilon, d, L, c, 1/c$ such that the activation function  of each neuron in the set of $\{\sigma, \sigma', ReLU\}$, and the following hold: If $\{\hat{x_t}\}_{t\in[0,1]}$ is the diffusion process governed by the It\^o SDE:
\begin{align}\label{SDE}
d\hat{x}_t = \hat{b}(\hat{x}_{t}, t)\dd t + \sqrt{2 } \dd W_t,
\end{align}
with $x_0 \sim p_1 = \law(y_1) \approx \gN(0, I)$ with the drift $\hat{b}(x,t) = - (x - 2 \hat{v}(x, 1-t))$, then $\hat{\mu} := \mathrm{Law}(\hat{x}_1)$, satisfies $D(\mu||\hat{\mu}) \leq \epsilon$.
\end{proposition}
\begin{remark}\label{rem:approx}
    Assuming $\pi$ satisfies a logarithmic Sobolev inequality, extending the time domain to $t\in [0,T]$ and sampling $\hat{x}_0 \sim \gN(0,I)$ approximately, it follows that $D(\mu||\hat{\mu}) \leq  e^{-T}\KL(\pi || \gN(0,1)) + T\epsilon$.
\end{remark}
The proof will closely follow \cite{tzen2019theoretical} however key steps must be slightly modified to show that the value function satisfies the required regularity properties to exploit the core results in \cite{tzen2019neural}.

\subsection{Prior Work - Heat Semigroup and F\"ollmer Drift}

Here we will introduce the heat semigroup and F\"ollmer drifts to highlight the previous work done in \citep{tzen2019theoretical}.
\begin{definition}
The heat semi-group is defined as 
\begin{align}
Q^{\sigma}_{t}f(y) &= \mathbb{E}_{Z \sim \mathcal{N}(0, I)}\left[f\left(y +\sigma t^{1/2} Z\right)\right],
\end{align}
and thus the F\"ollmer drift \citep{tzen2019theoretical} can be expressed as
\begin{align}
v^{*}_t(y) =  \nabla \ln Q^{\sigma}_{T-t}f(y).
\end{align}
\end{definition}
Where $v^{*}$ can be used to sample exactly from the desired target distribution by simulating the F\"{o}llmer drift SDE in \citep{tzen2019theoretical}, which coincides exactly with the backwards pinned Brownian motion  when setting $\alpha_t = \sigma^2 t$:\vspace{-0.5cm}\begin{align}
\mathrm{d}y_t &= \overbrace{\Big\{\frac{y_t}{t}+ \sigma^2 \nabla \log p_{T-t}(y_t)\Big\}}^{v^{*}_t(y_t)} \mathrm{d}t + \sigma \mathrm{d}W_t, \; y_0 =0. \nonumber
\end{align}
This process is commonly referred to as the Sch\"{o}dinger F\"{o}llmer. The prior seminal work of \cite{tzen2019theoretical} focuses on proving regularity properties of the heat semigroup as well as expressiveness remarks for the  F\"{o}llmer  drift. In this work, we port over these results to denoising diffusion models (i.e. VP-SDE based models).


\subsection{OU Semigroup and Time Reversal}\label{ousec}

This section introduces the OU semigroup \citep{metafune2002spectrum} whose logarithmic gradient can be directly connected to the score \citep{song2020score} in Equation \ref{eq:forwarddiffusionP}. Based on this reformulation of the score we can extend the results from \cite{tzen2019theoretical} to denoising diffusion via VP-SDEs. In the remainder of this section, we will introduce new results pertaining to the regularity properties of this operator that will enable us to prove Proposition  \ref{col:est}.

\begin{definition}
We define the VP-SDE semigroup as,
\begin{align}
U^{\beta_t}_{t}f(y) &= \mathbb{E}_{Z \sim \mathcal{N}(0, I)}\left[f\left(e^{-\int_0^{t} \beta_s \mathrm{d} s }y \right.\right. \nonumber \\
&\quad \left.\left.+\sigma(1-e^{-2\int_0^{t} \beta_s \mathrm{d}s})^{1/2} Z\right)\right].
\end{align}
Then the OU-semigroup \citep{metafune2002spectrum} (typically defined with $\beta_t=\beta=1$) is a simpler instance of the above
\begin{align}
U^{\beta}_{t}f(y) &= \mathbb{E}_{Z \sim \mathcal{N}(0, I)}\left[f\left(e^{- \beta t }y +\sigma(1-e^{-2 \beta t})^{1/2} Z\right)\right]. \nonumber
\end{align}
\end{definition}

For simplicity we will be working with the OU semi-group when $\beta=1$ (denoted $U_{t}$), however, these results can be extended to the more general case. In the following remark, we highlight the connection between the OU semi-group, the value function, and the score in DDPM.

 \begin{remark}\label{rem:ou}
The drift of the time reversal of the VP-SDE (i.e. $b^{*}(y,t)= -\beta_{T-t} (y -2\sigma^2 \nabla \log \phi_{T-t}(y))$) can be expressed in terms of the OU semigroup via:
\begin{align}
    \nabla \log \phi_{T-t}(y) = \nabla_y \ln U_{T-t}^{\beta_{t}}f(y).
\end{align}
 When $f(x) = \frac{\mathrm{d}\pi}{\mathrm{d}\gN(0, \sigma^2 I)}(x)$. This in turn can be related to the score 
\begin{multline}
\nabla \log p_{T-t}(y) = -\left(\frac{y}{2\sigma^2}-\nabla \log \phi_{T-t}(y)\right) \\
= -\left(\frac{y}{2\sigma^2}-\nabla_y \ln U_{T-t}^{\beta_{t}}f(y)\right).
\end{multline}

\end{remark}
 
From this stage on we consider the case where $\sigma = \beta = 1$. Notice how the formulation in Remark \ref{rem:ou} is reminiscent of the F\"ollmer drift \citep{follmer1984entropy,dai1991stochastic,tzen2019theoretical,huang2021schrodinger}. Finally, we highlight that it is this very simple remark that facilitates porting over the general proof strategy from \citep{tzen2019theoretical} to diffusion-based models. Furthermore, we remind the reader that the results in \cite{tzen2019theoretical} only apply to the F\"ollmer drift and the heat semigroup (i.e.$\nabla_y \ln \phi_t(y)=  \nabla_y \ln Q_{t}f(y)$ with $Q_{t}f(y)  = \mathbb{E}_{Z \sim \mathcal{N}(0, I)}\left[f\left(y+\sqrt{t} Z\right)\right] $), {thus requiring new results. 

\subsection{Regularity Properties }\label{sec:regularity}
In this section, we will prove regularity properties pertaining to the OU semigroup which will allow us to extend the theoretical guarantees in \cite{tzen2019theoretical} to denoising diffusion models and samplers \citep{song2020score, ho2020denoising,vargas2023denoising}. Moving forward we prove a basic auxiliary result regarding the commutativity of the OU-semigroup with partial derivatives. From this result, by using Corollary \ref{reg:corr}, we could bound the OU-semigroup norm when differentiated. Proofs for the following results can be found in  Appendix \ref{reg}, and Appendix \ref{covering}.

\begin{lemma}\label{lem:ou_commute}
OU semigroup is commutative with the gradient operator that is for $f:\sR^d \to \sR$ we have $\partial_{y_i } U_t f(y) = U_t \partial_{y_i} f(y)$.
\end{lemma}

\subsubsection{Terminal Cost} \label{sec:terminal_cost}

Contrary to \cite{tzen2019neural}  $g_{x,t}(z) = g(e^{-t}x + (1-e^{-2t})^{1/2}z)$ (where $x\in B^d(R), z\in \sR^d$)is not Lipschitz in a Euclidean sense. As a result the standard Euclidean covering-number properties used in \cite{tzen2019neural} no longer apply and thus we have to derive bounds for these quantities from scratch.

Across this section will refer to $g_{x,t}(z)$ as the terminal cost, due to its role in stochastic control. We want to underline to the reader that this quantity is of high importance as the optimal drift can be expressed in terms of the OU-semigroup is applied to the terminal cost ($\nabla \ln \phi_t(x) = \nabla \ln U_t g_{x,t}(z)$) when $g=f$. 

\begin{itemize}
    \item First we prove that a centered version of the terminal cost is $\mathscr{L}^2(Q)$ Lipchitz with respect to a newly defined metric. This will allow us to obtain a bound for the covering number of a function class induced by the terminal cost.
    \item We then derive an envelope for the terminal cost. This in conjunction with further results on covering numbers allows us to control Dudley's entropy integral \citep{dudley1967sizes}. This in turn enables results from empirical process theory \citep{gine2021mathematical} that quantify the error for an empirical estimate of the OU semigroup.
\end{itemize}


\begin{figure*}[t]
  \centering
  \begin{subfigure}{\textwidth}
    \centering
    \includegraphics[width=\linewidth]{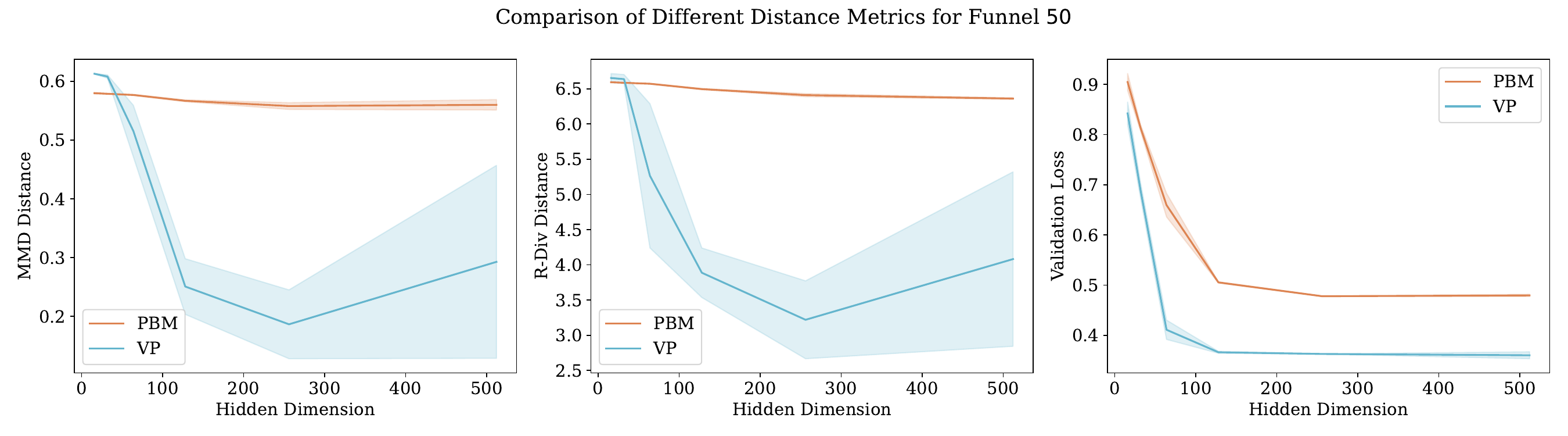}
    \label{fig:funnel_50}
  \end{subfigure}

  \begin{subfigure}{\textwidth}
    \centering
    \includegraphics[width=\linewidth]{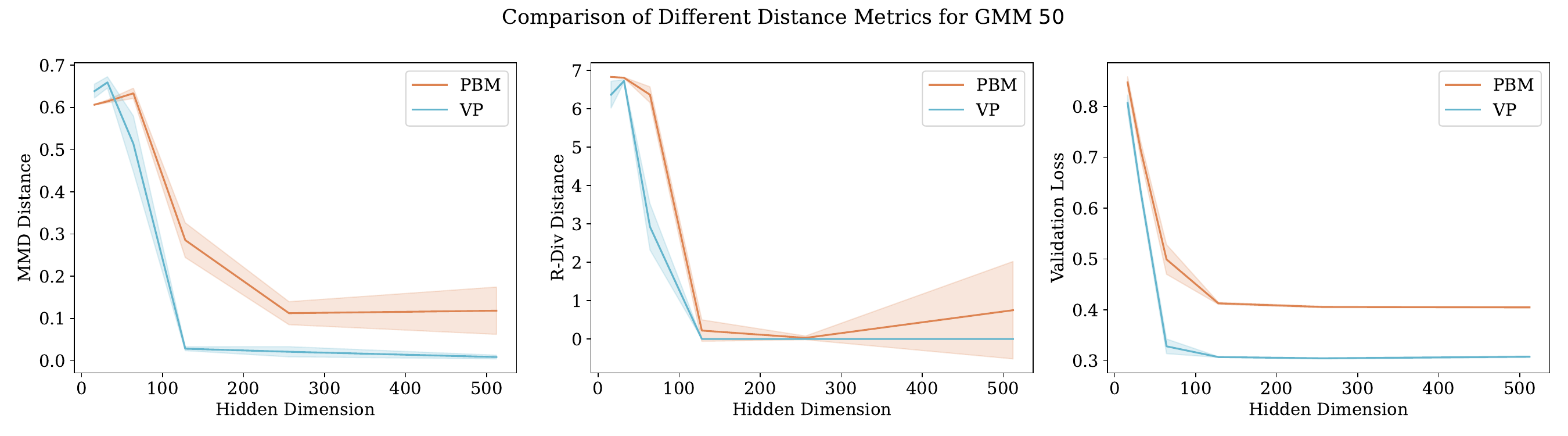}
    \label{fig:gmm_50}
  \end{subfigure}

  \caption{Comparison of distances between $\pi$ and $p^{\mathrm{model}}_\theta$ at time $T$ over $50$-dimensional Funnel and GMM-10 datasets.}
  \label{fig:combined_hid_dim}
\end{figure*}

\begin{lemma}\label{lem:expected_lip} ($\mathscr{L}^2$ Lipchitz condition)
    Let $\bar{g}_{t,x}(z) = g(e^{-t}x + (1-e^{-2t})^{1/2}z) - g(0)$ then it follows that:
\begin{align*}
|| \bar{g}_{t,x}(z) - \bar{g}_{t',x'}(z)||_{\mathscr{L}^2(Q)} \leq \\
 L\left(1 + \sqrt{2}||z||_{\mathscr{L}^2(Q)} \right) \rho_{OU}((t,x), (t',x')), 
\end{align*}
such that $\rho_{OU}((t,x), (t',x'))\!\! =\!\! || e^{-t}x - x'e^{-t'}||  + |t - t|^{1/2}$.
\end{lemma}
\begin{lemma}\label{lem:envelope} 
    Let $g : \mathbb{R} ^d \to \mathbb{R} $ be L-Lipschitz with respect to the Euclidean norm. Then for $F(z) := L((R \vee 1) + \sqrt{2}||z||)$ we have: 
\begin{align}
    \Big|g\left(e^{- t}x+(1-e^{-2 t})^{1/2} z\right) - g(0)\Big| \leq F(z).
\end{align}
\end{lemma}




    

\subsubsection{Covering Number}\label{covering_number}

We must first obtain a notion of approximation "difficulty" in approximating the OU-semigroup to obtain estimation errors on the score. To do so, we must obtain bounds quantifying how many "tiles" (closed balls) are required to cover the space $\gG$. The formal mechanism to do so is known as a covering number.

The $\mathscr{L}^2(Q)$ covering number of the function space $\gG$ is defined by:
\begin{align*}
N\left(\mathcal{G}, \mathscr{L}^2(Q), \varepsilon\right):=\min \left\{K: \exists f_1, \ldots, \exists f_K \in \mathscr{L}^2(Q) \right. \\
\text { s.t. } \left. \sup _{q \in \mathcal{G}} \min _{k \leq K}\left\|g-f_k\right\|_{L^2(P)} \leq \varepsilon\right\}.
\end{align*}
In general, the covering number $N\left(\gA, \rho, \varepsilon\right)$ is the smallest number of balls of size $\epsilon$ wrt to the metric $\rho$ that cover the set $\gA$. Once we obtain the appropriate bound on $N\left(\mathcal{G}, \mathscr{L}^2(Q), \varepsilon\right)$ the results from \citep{tzen2019theoretical} follow with minor modifications and thus Corollary \ref{col:est} will follow. In this section we will be bounding the $\mathscr{L}^2(Q)$ covering number of the function space $\mathcal{G}:=\left\{\bar{g}_{x, t}: x \in \mathrm{B}^d(R), t \in[0,1]\right\}$.
\begin{lemma}\label{lem:metlip}
Given the metric space $\big( [0,T] \times B^d(R) , \rho_{OU}\big)$ where:
\begin{align}
    \rho_{OU}((t,x), (t',x')) = || e^{-t}x - x'e^{-t'}||  + |t - t'|^{1/2}, \nonumber
\end{align}
and $||(t,x)||_{OU} =  \rho_{OU}((t,x), (0, 0))= || e^{-t}x ||  + |t|^{1/2}$.
It follows that:
\begin{align}
     N(\gG,  \mathscr{L}^2(Q), \epsilon ||F ||_{\mathscr{L}^2(Q)}) \leq  N([0,T] \times B^d(R),  \rho_{OU}, \epsilon).  \nonumber
\end{align}
\end{lemma}


\begin{lemma}\label{lem:covprod}
Given the metric space $\big( [0,T] \times B^d(R) , \rho_{OU}\big)$ it follows that:
\begin{multline}
N([0,T] \times B^d(R),  \rho_{OU}, \epsilon)  \leq \\
N([0,T], |\cdot|,  \epsilon^2/4) N(B^d(R), ||\cdot ||, \epsilon/2).
\end{multline}
\end{lemma}

From Lemmas \ref{lem:metlip}, \ref{lem:covprod} it follows that:
\begin{multline}
N(\gG,  \mathscr{L}^2(Q), \epsilon ||F ||_{\mathscr{L}^2(Q)}) \leq \\
N([0,T], |\cdot|,  \epsilon^2/4) N(B^d(R), ||\cdot ||, \epsilon/2)
\end{multline}
Establishing the existence of such bounds will facilitate our ability to subsequently demonstrate the approximation results. We will move forward in presenting two tighter bounds for the error. Proofs for Lemmas \ref{lem:metlip} and \ref{lem:covprod} can be found in Appendix \ref{covering}. 

\subsubsection{Sharper Bounds for {OU Semigroup} Covers} \label{sec:shaper}

In this section, we will present two bounds concerning the metric space $\big( [0,T] \times B^d(R), \rho_{OU}\big)$. The first bound is an extension of the heat semigroup results presented in \cite{tzen2019theoretical} in $[0,1]$ to $[0,T]$:

\begin{corollary}\label{lem:first_bound}

    Given the metric space $\big( [0,T] \times B^d(R) , \rho_{OU}\big)$ it follows that:
    \begin{align}
         N(B^d(R), ||\cdot ||, \epsilon/2) N([0,T], |\cdot|,  \epsilon^2/4)\leq \left(\frac{2\sqrt{3R T}}{\epsilon}\right)^{2d}. \nonumber
    \end{align}
\end{corollary}

The second bound is obtained from the properties of the OU process, establishing a tighter bound for this particular metric space.

\begin{proposition}\label{prop:better_bound}
    Given the metric space $\big( [0,T] \times B^d(R) , \rho_{OU}\big)$ it follows that:
    \begin{align}
         N([0,T] \times B^d(R),  \rho_{OU}, \epsilon)  \leq \left( \frac{ 2   e^{-\epsilon^2/2}  \sqrt{3TR}}{\epsilon}\right)^{d}. \nonumber
    \end{align}
\end{proposition}



\begin{figure*}[t]
  \centering
  \begin{subfigure}{\textwidth}
    \centering
    \includegraphics[width=\linewidth]{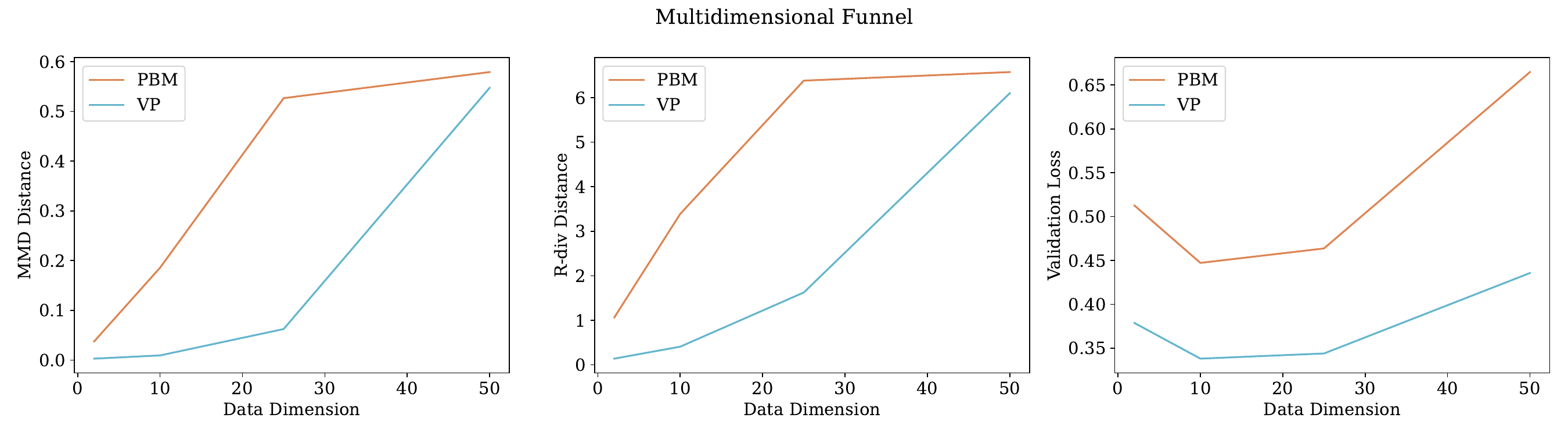}
    \label{fig:funnel_64}
  \end{subfigure}
  \begin{subfigure}{\textwidth}
    \centering
    \includegraphics[width=\linewidth]{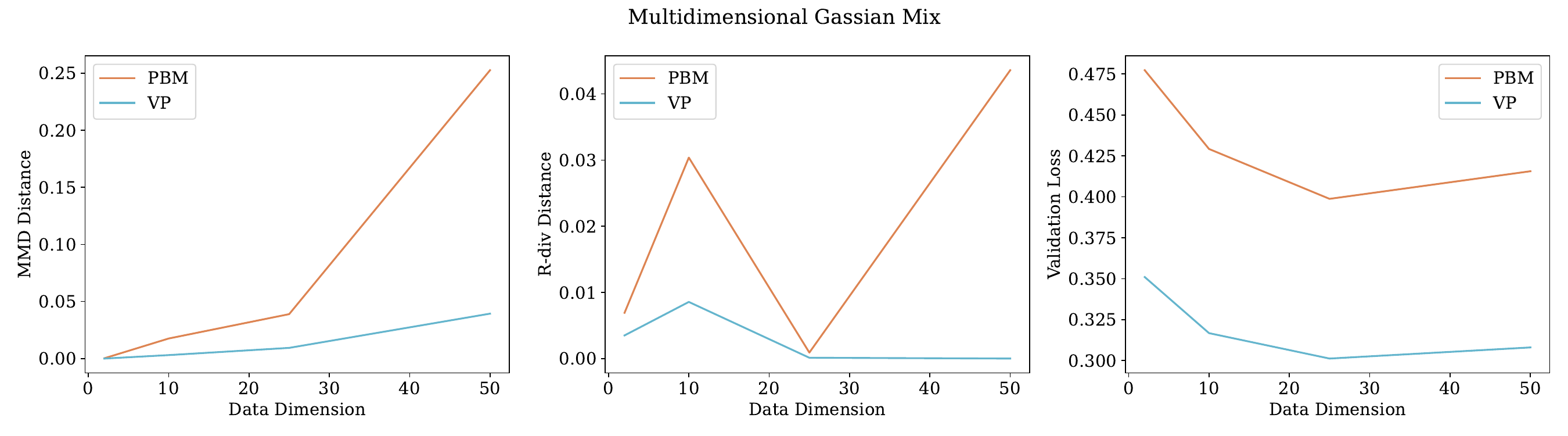}
    \label{fig:gmm_dimension_64}
  \end{subfigure}

  \caption{Comparison of distances between $\pi$ and $p^{\mathrm{model}}_\theta$ at time $T$ over Funnel and Mixed Gaussians varying in dimensions.}
  \label{fig:combined_dim}
\end{figure*}

\subsection{Score Estimation Results}

The following result constitutes an important piece in proving Proposition \ref{col:est}, since the way $N$ is picked depends on the previously mentioned Lemmas.

\begin{corollary}\label{col:tzen} For any $\varepsilon>0$ and any $R>0$, there exist ${N}=\operatorname{poly}(1 / \varepsilon, d, L, R, T )$ points $z_1, \ldots, z_{{N}} \in \mathbb{R}^d$, and for $p(x,t,z_n) = e^{- t}x+(1-e^{-2 t})^{1/2} z_n$, for which the following holds:
\begin{align*}
\max _{n \leq N}\left\|z_n\right\| &\leq 8 \sqrt{(d+6) \log {N}},\\
\sup _{x \in \mathrm{B}^d(R)} &\sup _{t \in[0,1]}\left| \sum_{n=1}^{{N}} {\frac{\nabla {f\left(p(x,t,z_n)\right)}}{N}}-U_t f(x)\right| \leq \varepsilon,\\
\sup _{x \in \mathrm{B}^d(R)} &\sup _{t \in[0,1]}\left\| \sum_{n=1}^{{N}} \frac{\nabla {f\left(p(x,t,z_n)\right)}}{N}-\nabla U_t f(x)\right\| \leq \varepsilon.
\end{align*}
\end{corollary}

Following the steps from \cite{tzen2019theoretical}, Lemma \ref{lem:first_bound} and Corollary \ref{col:tzen}, we arrive at one of the main results that will help prove Proposition \ref{col:est}.

Fortunately, the first bound derived in Lemma \ref{lem:first_bound} arrives at a computable integral over the \textit{Koltchinskii-Pollard} $\epsilon$\textit{-entropy}, which is needed for the completion of the proof of Corollary \ref{col:tzen}, which then is used in proving Proposition \ref{col:est} (see Appendix \ref{apdx:approx}). However, for the second bound, despite being tighter, the integral is not tractable.  A more detailed description of this can be found in the Appendix proof of Lemma D.2, and the following remark.

Following results in Proposition \ref{lem:first_bound} and Corollary \ref{prop:better_bound} we can observe the presence of a $e^{-\frac{d\eps^2}{2}}$ term in the cover for the OU semi-group, thus its cover is smaller than that of the heat semigroup, this motivates the following observation:

\begin{observation}\label{obs:tight_bound}
The \textit{Koltchinskii-Pollard} $\epsilon$\textit{-entropy} that corresponds to the OU semigroup lower bounds that of the heat semi-group.
\end{observation}
We believe Observation \ref{obs:tight_bound} motivates how the score for VP-SDEs forms a simpler function class than the score for the PBM, which potentially indicates that the score on VP-SDE admits a neural network estimator achieving a smaller error than that of the score of a PBM \citep{tzen2019theoretical}, we will now explore this conjecture empirically.

\section{Simulations}\label{sec:simulation}

In this section, we evaluate the performance of VP-SDE and PBM using the score matching loss \citep{song2020score} together with other metrics across different network sizes and dataset dimensions. Section \ref{sec:syntethic} analyzes performance on two synthetic datasets, and Section \ref{sec:image} compares the performance on image data. Detailed information on learning rates, dataset splits, training epochs, and noise schedules is provided in Appendices \ref{appdx:sim} and \ref{app:image_data_details}, respectively.

\subsection{Syntethic Simulations}
\label{sec:syntethic}
We explore both VP-SDE and PBM across two simulated datasets, selected due to their flexibility in being able to increase the dimension of space.

\textbf{GMM-10:} We use a Gaussian mixture model with $10$ mixtures, each mixture is parameterized $\mathcal{N}(\mu_i, I)$ where $\mu_i$ is sampled uniformly within the $(d-1)$-ball of Radius $6$. 

\textbf{Neals Funnel \citep{neal2011mcmc}}  This d-dimensional challenging distribution is given by $\gamma(x_{1:d})= \mathcal{N}(x_1;0, \sigma_{f}^2)\mathcal{N}(x_{2:d};0, \exp(x_1) I)$,
where $\sigma_{f}^2=9$.
\subsubsection{Evaluation Metrics}

To assess which method we report the following performance metrics across a series of numerical simulations:

\textbf{Score matching loss}: We report the score matching loss \citep{song2020score} on a hold test set. This loss acts as a proxy to measure how well the trained network has learned the score. 

\textbf{MMD}: We the use maximum mean discrepancy metric \citep{gretton2012kernel} to measure the distance between $\mathcal{D}(p^{\mathrm{model}}_\theta, \pi)$. The motivation for this is that the KL-divergence between the marginals $p^{\mathrm{model}}_\theta$ and $\pi$ (via data processing and Girsanov Theorem) is upper bounded by the error between the scoring network and the true score, thus a better performance in score matching typically indicates better marginal performance, here we assess the latter.

\textbf{r-divergence}: Similar to the MMD experiments we explore an additional divergence (the r-divergence \citep{zhao2023r}) to more thoroughly verify the performance in sampling the data distribution.

Note for the MMD and r-divergence metrics we use $1000$ samples from the trained score models and the target distributions to compute the aforementioned metrics and $20000$ samples for the validation score matching loss. More details can be found in Appendix \ref{appdx:sim}.

\subsubsection{Score Estimation Across Network Widths}\label{sec:increase_hid_dim}

For this experiment, we fix the dimensions of the data sets to $d=50$ and vary the width of the score networks across $4,16,32,64,128,256,512$. From our results, we can see that in Figure \ref{fig:combined_hid_dim} on the left-hand side, VP-SDE attains a lower score-matching loss for the same number of parameters and can sample the target distribution better than PBM, suggesting that VP-SDE requires less expressive networks to be estimated, which is in agreement with the insights we obtained from our covering number results.

In Figure \ref{fig:combined_hid_dim}, for the Funnel dataset, VP seems to express a double descent \cite{nakkiran2021deep, d2020double} type of behavior. As the hidden layer dimension passes $256$ parameters, the model generates samples that are further away from $\pi$. 

We also ran experiments for $d=10$, and the same behavior can be noticed across all network widths. The experiments can be viewed in Appendix \ref{appx:dim_10}. In this case, for the Funnel dataset, the double descent behavior can be noticed in both VP and PBM cases, for MMD and R-Div metrics. 

\begin{figure}[t]
    \centering
    \includegraphics[width=\linewidth]{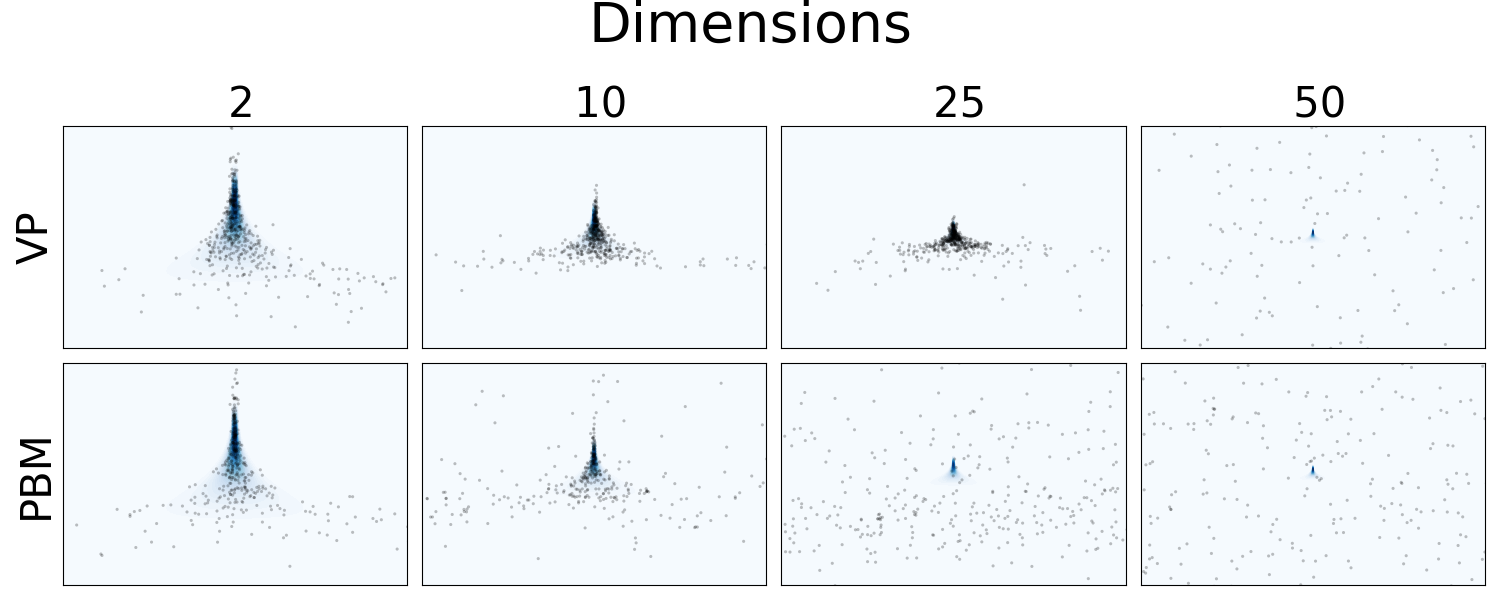}
    \caption{Samples (2D slice $(x_1, x_0)$) from PBM and VP trained on various sizes of the Funnel distribution. In the background probability density of the Funnel distribution.}
    \label{fig:evolution}
\end{figure}

\subsubsection{Score Estimation Across Data-set Dimensions}

In this sequence of experiments depicted in Figure \ref{fig:combined_dim} on the right, we maintain a fixed network width of $64$ while varying the dimensionality across $2, 10, 25, 50$ to evaluate the performance of both methods in estimating the score as the dimensionality of the target samples increases. For dimensions below $50$, VP tends to sample points that match the target/data distribution more closely on the Funnel dataset, whereas, at $50$ dimensions, both their performances start to degrade similarly (e.g. also see  Figure \ref{fig:evolution}). Finally, VP consistently produces superior samples in the case of the GMM targets, thus corroborating our observation.

\subsection{Image Data Simultation}
\label{sec:image}
In this section, we investigate the performance of VP-SDE and PBM on image datasets, specifically MNIST, MNIST-Fashion, and CIFAR-10. We conducted all experiments using a simple single-block ResNet architecture with three feature maps with a total number of $1.34$ million parameters (only $3\%$ of the number of the parameters used in \cite{song2020denoising}). Given the nature of image datasets, we evaluated the quality of generated samples using the Frechet Inception Distance (FID) \cite{jiralerspong2023feature}. The nature of the experiment is to see the difference in the relative performance of the two forward processes. Further details, and generated samples are provided in Appendix \ref{app:image_data_details}.

\begin{table}[htbp]
\centering
\caption{Comparison of FID scores on the test set of MNIST, Fashion MNIST, and CIFAR-10 datasets}
\label{tab:fid_comparison}
\begin{adjustbox}{width=\linewidth}
\begin{tabular}{@{}lccc@{}}
\toprule
Dataset & MNIST & Fashion MNIST & CIFAR-10 \\ \midrule
VP-SDE & $13.86 \pm 0.06$ & $6.99 \pm 0.03$ &  $40.09 \pm 0.88$  \\
PBM & $15.59 \pm 0.07$ & $17.26 \pm 0.35$ & $49.47 \pm 0.67$ \\ \bottomrule
\end{tabular}
\end{adjustbox}
\end{table}

As shown in Table \ref{tab:fid_comparison}, given the same computation/parameter budget, VP achieves better estimates than PBM across all image datasets, with a notable improvement in performance observed in MNIST-Fashion. This suggests that the balance between the complexity of the datasets and the size of the network was optimized for this particular dataset.




\section{Conclusion}

We establish a connection between the VP-SDE score and the OU-semigroup, revealing similarities between F\"{o}llmer drift-based and DDPM-based sampling approaches. Using this connection, we demonstrate how the VP-SDE score can be approximated efficiently by multilayer neural networks, under fairly general assumptions on the target distribution. To exploit previous results on the F\"{o}llmer drift \citep{tzen2019theoretical} we establish novel regularity properties for the OU-semigroup that allow us to adapt the results in \cite{tzen2019theoretical} to our setting. Finally motivated by our theoretical results we empirically demonstrate how a VP-SDE based forward process can be approximated better by a neural network of the same size than one with a PBM-SDE forward process.



\nocite{langley00}

\bibliography{bib}

\begin{thebibliography}{8}
\providecommand{\natexlab}[1]{#1}
\providecommand{\url}[1]{\texttt{#1}}
\expandafter\ifx\csname urlstyle\endcsname\relax
  \providecommand{\doi}[1]{doi: #1}\else
  \providecommand{\doi}{doi: \begingroup \urlstyle{rm}\Url}\fi

\bibitem[Author(2021)]{anonymous}
Author, N.~N.
\newblock Suppressed for anonymity, 2021.

\bibitem[Duda et~al.(2000)Duda, Hart, and Stork]{DudaHart2nd}
Duda, R.~O., Hart, P.~E., and Stork, D.~G.
\newblock \emph{Pattern Classification}.
\newblock John Wiley and Sons, 2nd edition, 2000.

\bibitem[Kearns(1989)]{kearns89}
Kearns, M.~J.
\newblock \emph{Computational Complexity of Machine Learning}.
\newblock PhD thesis, Department of Computer Science, Harvard University, 1989.

\bibitem[Langley(2000)]{langley00}
Langley, P.
\newblock Crafting papers on machine learning.
\newblock In Langley, P. (ed.), \emph{Proceedings of the 17th International
  Conference on Machine Learning (ICML 2000)}, pp.\  1207--1216, Stanford, CA,
  2000. Morgan Kaufmann.

\bibitem[Michalski et~al.(1983)Michalski, Carbonell, and
  Mitchell]{MachineLearningI}
Michalski, R.~S., Carbonell, J.~G., and Mitchell, T.~M. (eds.).
\newblock \emph{Machine Learning: An Artificial Intelligence Approach, Vol. I}.
\newblock Tioga, Palo Alto, CA, 1983.

\bibitem[Mitchell(1980)]{mitchell80}
Mitchell, T.~M.
\newblock The need for biases in learning generalizations.
\newblock Technical report, Computer Science Department, Rutgers University,
  New Brunswick, MA, 1980.

\bibitem[Newell \& Rosenbloom(1981)Newell and Rosenbloom]{Newell81}
Newell, A. and Rosenbloom, P.~S.
\newblock Mechanisms of skill acquisition and the law of practice.
\newblock In Anderson, J.~R. (ed.), \emph{Cognitive Skills and Their
  Acquisition}, chapter~1, pp.\  1--51. Lawrence Erlbaum Associates, Inc.,
  Hillsdale, NJ, 1981.

\bibitem[Samuel(1959)]{Samuel59}
Samuel, A.~L.
\newblock Some studies in machine learning using the game of checkers.
\newblock \emph{IBM Journal of Research and Development}, 3\penalty0
  (3):\penalty0 211--229, 1959.

\end{thebibliography}


\begin{thebibliography}{54}
\providecommand{\natexlab}[1]{#1}
\providecommand{\url}[1]{\texttt{#1}}
\expandafter\ifx\csname urlstyle\endcsname\relax
  \providecommand{\doi}[1]{doi: #1}\else
  \providecommand{\doi}{doi: \begingroup \urlstyle{rm}\Url}\fi

\bibitem[Bakry et~al.(2014)Bakry, Gentil, Ledoux, et~al.]{bakry2014analysis}
Dominique Bakry, Ivan Gentil, Michel Ledoux, et~al.
\newblock \emph{Analysis and geometry of Markov diffusion operators}, volume
  103.
\newblock Springer, 2014.

\bibitem[Berner et~al.(2022)Berner, Richter, and Ullrich]{berner2022optimal}
Julius Berner, Lorenz Richter, and Karen Ullrich.
\newblock An optimal control perspective on diffusion-based generative
  modeling.
\newblock \emph{arXiv preprint arXiv:2211.01364}, 2022.

\bibitem[Bernton et~al.(2019)Bernton, Heng, Doucet, and
  Jacob]{bernton2019SBsamplers}
Espen Bernton, Jeremy Heng, Arnaud Doucet, and Pierre~E Jacob.
\newblock Schr\"odinger bridge samplers.
\newblock \emph{arXiv preprint arXiv:1912.13170}, 2019.

\bibitem[Biggs et~al.(2023)Biggs, Schrab, and Gretton]{biggs2023mmd}
Felix Biggs, Antonin Schrab, and Arthur Gretton.
\newblock Mmd-fuse: Learning and combining kernels for two-sample testing
  without data splitting.
\newblock \emph{arXiv preprint arXiv:2306.08777}, 2023.

\bibitem[Blei et~al.(2017)Blei, Kucukelbir, and McAuliffe]{blei2017variational}
David~M Blei, Alp Kucukelbir, and Jon~D McAuliffe.
\newblock Variational inference: A review for statisticians.
\newblock \emph{Journal of the American statistical Association}, 112\penalty0
  (518):\penalty0 859--877, 2017.

\bibitem[Bubeck et~al.(2018)Bubeck, Eldan, and Lehec]{bubeck2018sampling}
S{\'e}bastien Bubeck, Ronen Eldan, and Joseph Lehec.
\newblock Sampling from a log-concave distribution with projected langevin
  monte carlo.
\newblock \emph{Discrete \& Computational Geometry}, 59:\penalty0 757--783,
  2018.

\bibitem[Chen et~al.(2023)Chen, Huang, Zhao, and Wang]{chen2023score}
Minshuo Chen, Kaixuan Huang, Tuo Zhao, and Mengdi Wang.
\newblock Score approximation, estimation and distribution recovery of
  diffusion models on low-dimensional data.
\newblock In \emph{International Conference on Machine Learning}, pages
  4672--4712. PMLR, 2023.

\bibitem[Chen et~al.(2022)Chen, Chewi, Li, Li, Salim, and
  Zhang]{chen2022sampling}
Sitan Chen, Sinho Chewi, Jerry Li, Yuanzhi Li, Adil Salim, and Anru~R Zhang.
\newblock Sampling is as easy as learning the score: theory for diffusion
  models with minimal data assumptions.
\newblock \emph{arXiv preprint arXiv:2209.11215}, 2022.

\bibitem[Cole and Lu(2024)]{cole2024score}
Frank Cole and Yulong Lu.
\newblock Score-based generative models break the curse of dimensionality in
  learning a family of sub-gaussian probability distributions.
\newblock \emph{arXiv preprint arXiv:2402.08082}, 2024.

\bibitem[Dai~Pra(1991)]{dai1991stochastic}
Paolo Dai~Pra.
\newblock A stochastic control approach to reciprocal diffusion processes.
\newblock \emph{Applied Mathematics and Optimization}, 23\penalty0
  (1):\penalty0 313--329, 1991.

\bibitem[De~Bortoli et~al.(2021)De~Bortoli, Thornton, Heng, and
  Doucet]{debortoli2021diffusion}
Valentin De~Bortoli, James Thornton, Jeremy Heng, and Arnaud Doucet.
\newblock Diffusion {S}chr{\"o}dinger bridge with applications to score-based
  generative modeling.
\newblock In \emph{Advances in Neural Information Processing Systems}, 2021.

\bibitem[Doucet et~al.(2022)Doucet, Grathwohl, Matthews, and
  Strathmann]{doucet2022annealed}
Arnaud Doucet, Will Grathwohl, Alexander G de~G Matthews, and Heiko Strathmann.
\newblock Score-based diffusion meets annealed importance sampling.
\newblock In \emph{Advances in Neural Information Processing Systems}, 2022.

\bibitem[Dudley(1967)]{dudley1967sizes}
Richard~M Dudley.
\newblock The sizes of compact subsets of hilbert space and continuity of
  gaussian processes.
\newblock \emph{Journal of Functional Analysis}, 1\penalty0 (3):\penalty0
  290--330, 1967.

\bibitem[d’Ascoli et~al.(2020)d’Ascoli, Refinetti, Biroli, and
  Krzakala]{d2020double}
St{\'e}phane d’Ascoli, Maria Refinetti, Giulio Biroli, and Florent Krzakala.
\newblock Double trouble in double descent: Bias and variance (s) in the lazy
  regime.
\newblock In \emph{International Conference on Machine Learning}, pages
  2280--2290. PMLR, 2020.

\bibitem[Fedus et~al.(2018)Fedus, Rosca, Lakshminarayanan, Dai, Mohamed, and
  Goodfellow]{many-paths}
W.~Fedus, M.~Rosca, B.~Lakshminarayanan, A.~M. Dai, S.~Mohamed, and
  I.~Goodfellow.
\newblock Many paths to equilibrium: {GAN}s do not need to decrease a
  divergence at every step.
\newblock In \emph{ICLR}, 2018.

\bibitem[Fleming and Rishel(2012)]{fleming2012deterministic}
Wendell~H Fleming and Raymond~W Rishel.
\newblock \emph{Deterministic and stochastic optimal control}, volume~1.
\newblock Springer Science \& Business Media, 2012.

\bibitem[F\"ollmer(1984)]{follmer1984entropy}
Hans F\"ollmer.
\newblock An entropy approach to the time reversal of diffusion processes.
\newblock \emph{Lecture Notes in Control and Information Sciences},
  69:\penalty0 156--163, 1984.

\bibitem[Gin{\'e} and Nickl(2021)]{gine2021mathematical}
Evarist Gin{\'e} and Richard Nickl.
\newblock \emph{Mathematical foundations of infinite-dimensional statistical
  models}.
\newblock Cambridge university press, 2021.

\bibitem[Gretton et~al.(2012)Gretton, Borgwardt, Rasch, Sch{\"o}lkopf, and
  Smola]{gretton2012kernel}
Arthur Gretton, Karsten~M Borgwardt, Malte~J Rasch, Bernhard Sch{\"o}lkopf, and
  Alexander Smola.
\newblock A kernel two-sample test.
\newblock \emph{Journal of Machine Learning Research}, 2012.

\bibitem[Haussmann and Pardoux(1986)]{haussmann1986time}
Ulrich~G Haussmann and Etienne Pardoux.
\newblock Time reversal of diffusions.
\newblock \emph{The Annals of Probability}, 14\penalty0 (3):\penalty0
  1188--1205, 1986.

\bibitem[Ho et~al.(2020)Ho, Jain, and Abbeel]{ho2020denoising}
Jonathan Ho, Ajay Jain, and Pieter Abbeel.
\newblock Denoising diffusion probabilistic models.
\newblock In \emph{Advances in Neural Information Processing Systems}, 2020.

\bibitem[Huang et~al.(2021{\natexlab{a}})Huang, Lim, and
  Courville]{huang2021variational}
Chin-Wei Huang, Jae~Hyun Lim, and Aaron Courville.
\newblock A variational perspective on diffusion-based generative models and
  score matching.
\newblock \emph{arXiv preprint arXiv:2106.02808}, 2021{\natexlab{a}}.

\bibitem[Huang et~al.(2021{\natexlab{b}})Huang, Jiao, Kang, Liao, Liu, and
  Liu]{huang2021schrodinger}
Jian Huang, Yuling Jiao, Lican Kang, Xu~Liao, Jin Liu, and Yanyan Liu.
\newblock {S}chr{\"o}dinger-{F}{\"o}llmer sampler: {S}ampling without
  ergodicity.
\newblock \emph{arXiv preprint arXiv:2106.10880}, 2021{\natexlab{b}}.

\bibitem[Jiralerspong et~al.(2023)Jiralerspong, Bose, Gemp, Qin, Bachrach, and
  Gidel]{jiralerspong2023feature}
Marco Jiralerspong, Avishek~Joey Bose, Ian Gemp, Chongli Qin, Yoram Bachrach,
  and Gauthier Gidel.
\newblock Feature likelihood score: Evaluating generalization of generative
  models using samples, 2023.

\bibitem[Kappen et~al.(2012)Kappen, G{\'o}mez, and Opper]{kappen2012optimal}
Hilbert~J Kappen, Vicen{\c{c}} G{\'o}mez, and Manfred Opper.
\newblock Optimal control as a graphical model inference problem.
\newblock \emph{Machine learning}, 87\penalty0 (2):\penalty0 159--182, 2012.

\bibitem[Langley(2000)]{langley00}
P.~Langley.
\newblock Crafting papers on machine learning.
\newblock In Pat Langley, editor, \emph{Proceedings of the 17th International
  Conference on Machine Learning (ICML 2000)}, pages 1207--1216, Stanford, CA,
  2000. Morgan Kaufmann.

\bibitem[Lee et~al.(2023)Lee, Lu, and Tan]{lee2023convergence}
Holden Lee, Jianfeng Lu, and Yixin Tan.
\newblock Convergence of score-based generative modeling for general data
  distributions.
\newblock In \emph{International Conference on Algorithmic Learning Theory},
  pages 946--985. PMLR, 2023.

\bibitem[L{\'e}onard(2014)]{leonard2013survey}
Christian L{\'e}onard.
\newblock A survey of the {S}chr{\"o}dinger problem and some of its connections
  with optimal transport.
\newblock \emph{Discrete and Continuous Dynamical Systems-Series A},
  34\penalty0 (4):\penalty0 1533--1574, 2014.

\bibitem[Liu et~al.(2023)Liu, Vahdat, Huang, Theodorou, Nie, and
  Anandkumar]{liu20232}
Guan-Horng Liu, Arash Vahdat, De-An Huang, Evangelos~A Theodorou, Weili Nie,
  and Anima Anandkumar.
\newblock I $^{2}$ sb: Image-to-image schr\"{o}dinger bridge.
\newblock \emph{arXiv preprint arXiv:2302.05872}, 2023.

\bibitem[Ma et~al.(2019)Ma, Chen, Jin, Flammarion, and Jordan]{ma2019sampling}
Yi-An Ma, Yuansi Chen, Chi Jin, Nicolas Flammarion, and Michael~I Jordan.
\newblock Sampling can be faster than optimization.
\newblock \emph{Proceedings of the National Academy of Sciences}, 116\penalty0
  (42):\penalty0 20881--20885, 2019.

\bibitem[Metafune et~al.(2002)Metafune, Pallara, and
  Priola]{metafune2002spectrum}
Giorgio Metafune, Diego Pallara, and Enrico Priola.
\newblock Spectrum of ornstein-uhlenbeck operators in lp spaces with respect to
  invariant measures.
\newblock \emph{Journal of Functional Analysis}, 196\penalty0 (1):\penalty0
  40--60, 2002.

\bibitem[Nakkiran et~al.(2021)Nakkiran, Kaplun, Bansal, Yang, Barak, and
  Sutskever]{nakkiran2021deep}
Preetum Nakkiran, Gal Kaplun, Yamini Bansal, Tristan Yang, Boaz Barak, and Ilya
  Sutskever.
\newblock Deep double descent: Where bigger models and more data hurt.
\newblock \emph{Journal of Statistical Mechanics: Theory and Experiment},
  2021\penalty0 (12):\penalty0 124003, 2021.

\bibitem[Neal(2011)]{neal2011mcmc}
Radford~M Neal.
\newblock {MCMC} using {H}amiltonian dynamics.
\newblock \emph{Handbook of Markov Chain Monte Carlo}, 2\penalty0
  (11):\penalty0 2, 2011.

\bibitem[N{\"u}sken and Richter(2021)]{nusken2021solving}
Nikolas N{\"u}sken and Lorenz Richter.
\newblock Solving high-dimensional {H}amilton--{J}acobi--{B}ellman {PDE}s using
  neural networks: perspectives from the theory of controlled diffusions and
  measures on path space.
\newblock \emph{Partial Differential Equations and Applications}, 2\penalty0
  (4):\penalty0 1--48, 2021.

\bibitem[Oko et~al.(2023)Oko, Akiyama, and Suzuki]{oko2023diffusion}
Kazusato Oko, Shunta Akiyama, and Taiji Suzuki.
\newblock Diffusion models are minimax optimal distribution estimators.
\newblock In \emph{International Conference on Machine Learning}, pages
  26517--26582. PMLR, 2023.

\bibitem[Papamakarios et~al.(2021)Papamakarios, Nalisnick, Rezende, Mohamed,
  and Lakshminarayanan]{papamakarios2019normalizing}
George Papamakarios, Eric Nalisnick, Danilo~Jimenez Rezende, Shakir Mohamed,
  and Balaji Lakshminarayanan.
\newblock Normalizing flows for probabilistic modeling and inference.
\newblock \emph{The Journal of Machine Learning Research}, 22\penalty0
  (1):\penalty0 2617--2680, 2021.

\bibitem[Pham(2009)]{Pham:2009}
Huy{\^e}n Pham.
\newblock \emph{Continuous-time stochastic control and optimization with
  financial applications}, volume~61.
\newblock Springer Science \& Business Media, 2009.

\bibitem[Scott(1979)]{scott1979optimal}
David~W Scott.
\newblock On optimal and data-based histograms.
\newblock \emph{Biometrika}, 66\penalty0 (3):\penalty0 605--610, 1979.

\bibitem[Shalev-Shwartz and Ben-David(2014)]{shalev2014understanding}
Shai Shalev-Shwartz and Shai Ben-David.
\newblock \emph{Understanding machine learning: From theory to algorithms}.
\newblock Cambridge university press, 2014.

\bibitem[Sohl-Dickstein et~al.(2015)Sohl-Dickstein, Weiss, Maheswaranathan, and
  Ganguli]{sohl2015deep}
Jascha Sohl-Dickstein, Eric Weiss, Niru Maheswaranathan, and Surya Ganguli.
\newblock Deep unsupervised learning using nonequilibrium thermodynamics.
\newblock In \emph{International Conference on Machine Learning}, 2015.

\bibitem[Somnath et~al.(2023)Somnath, Pariset, Hsieh, Martinez, Krause, and
  Bunne]{somnath2023aligned}
Vignesh~Ram Somnath, Matteo Pariset, Ya-Ping Hsieh, Maria~Rodriguez Martinez,
  Andreas Krause, and Charlotte Bunne.
\newblock Aligned diffusion schr$\backslash$" odinger bridges.
\newblock \emph{arXiv preprint arXiv:2302.11419}, 2023.

\bibitem[Song et~al.(2020)Song, Meng, and Ermon]{song2020denoising}
Jiaming Song, Chenlin Meng, and Stefano Ermon.
\newblock Denoising diffusion implicit models.
\newblock \emph{arXiv preprint arXiv:2010.02502}, 2020.

\bibitem[Song et~al.(2021{\natexlab{a}})Song, Durkan, Murray, and
  Ermon]{song2021maximum}
Yang Song, Conor Durkan, Iain Murray, and Stefano Ermon.
\newblock Maximum likelihood training of score-based diffusion models.
\newblock In \emph{Advances in Neural Information Processing Systems},
  2021{\natexlab{a}}.

\bibitem[Song et~al.(2021{\natexlab{b}})Song, Sohl-Dickstein, Kingma, Kumar,
  Ermon, and Poole]{song2020score}
Yang Song, Jascha Sohl-Dickstein, Diederik~P Kingma, Abhishek Kumar, Stefano
  Ermon, and Ben Poole.
\newblock Score-based generative modeling through stochastic differential
  equations.
\newblock In \emph{International Conference on Learning Representations},
  2021{\natexlab{b}}.

\bibitem[Tzen and Raginsky(2019{\natexlab{a}})]{tzen2019neural}
Belinda Tzen and Maxim Raginsky.
\newblock Neural stochastic differential equations: Deep latent {G}aussian
  models in the diffusion limit.
\newblock \emph{arXiv preprint arXiv:1905.09883}, 2019{\natexlab{a}}.

\bibitem[Tzen and Raginsky(2019{\natexlab{b}})]{tzen2019theoretical}
Belinda Tzen and Maxim Raginsky.
\newblock Theoretical guarantees for sampling and inference in generative
  models with latent diffusions.
\newblock In \emph{Conference on Learning Theory}, 2019{\natexlab{b}}.

\bibitem[Vargas et~al.(2021{\natexlab{a}})Vargas, Ovsianas, Fernandes,
  Girolami, Lawrence, and N{\"u}sken]{vargas2021bayesian}
Francisco Vargas, Andrius Ovsianas, David Fernandes, Mark Girolami, Neil
  Lawrence, and Nikolas N{\"u}sken.
\newblock Bayesian learning via neural {S}chr{\"o}dinger--{F}{\"o}llmer flows.
\newblock \emph{arXiv preprint arXiv:2111.10510}, 2021{\natexlab{a}}.

\bibitem[Vargas et~al.(2021{\natexlab{b}})Vargas, Thodoroff, Lamacraft, and
  Lawrence]{vargasshro2021}
Francisco Vargas, Pierre Thodoroff, Austen Lamacraft, and Neil Lawrence.
\newblock Solving {S}chr{\"o}dinger bridges via maximum likelihood.
\newblock \emph{Entropy}, 23\penalty0 (9), 2021{\natexlab{b}}.
\newblock ISSN 1099-4300.
\newblock \doi{10.3390/e23091134}.
\newblock URL \url{https://www.mdpi.com/1099-4300/23/9/1134}.

\bibitem[Vargas et~al.(2023)Vargas, Grathwohl, and Doucet]{vargas2023denoising}
Francisco Vargas, Will~Sussman Grathwohl, and Arnaud Doucet.
\newblock Denoising diffusion samplers.
\newblock In \emph{The Eleventh International Conference on Learning
  Representations}, 2023.
\newblock URL \url{https://openreview.net/forum?id=8pvnfTAbu1f}.

\bibitem[Vempala and Wibisono(2019)]{vempala2019rapid}
Santosh Vempala and Andre Wibisono.
\newblock Rapid convergence of the unadjusted langevin algorithm: Isoperimetry
  suffices.
\newblock \emph{Advances in neural information processing systems}, 32, 2019.

\bibitem[Virtanen et~al.(2020)Virtanen, Gommers, Oliphant, Haberland, Reddy,
  Cournapeau, Burovski, Peterson, Weckesser, Bright, {van der Walt}, Brett,
  Wilson, Millman, Mayorov, Nelson, Jones, Kern, Larson, Carey, Polat, Feng,
  Moore, {VanderPlas}, Laxalde, Perktold, Cimrman, Henriksen, Quintero, Harris,
  Archibald, Ribeiro, Pedregosa, {van Mulbregt}, and {SciPy 1.0
  Contributors}]{2020SciPy}
Pauli Virtanen, Ralf Gommers, Travis~E. Oliphant, Matt Haberland, Tyler Reddy,
  David Cournapeau, Evgeni Burovski, Pearu Peterson, Warren Weckesser, Jonathan
  Bright, St{\'e}fan~J. {van der Walt}, Matthew Brett, Joshua Wilson, K.~Jarrod
  Millman, Nikolay Mayorov, Andrew R.~J. Nelson, Eric Jones, Robert Kern, Eric
  Larson, C~J Carey, {\.I}lhan Polat, Yu~Feng, Eric~W. Moore, Jake
  {VanderPlas}, Denis Laxalde, Josef Perktold, Robert Cimrman, Ian Henriksen,
  E.~A. Quintero, Charles~R. Harris, Anne~M. Archibald, Ant{\^o}nio~H. Ribeiro,
  Fabian Pedregosa, Paul {van Mulbregt}, and {SciPy 1.0 Contributors}.
\newblock {{SciPy} 1.0: Fundamental Algorithms for Scientific Computing in
  Python}.
\newblock \emph{Nature Methods}, 17:\penalty0 261--272, 2020.
\newblock \doi{10.1038/s41592-019-0686-2}.

\bibitem[Wainwright et~al.(2008)Wainwright, Jordan,
  et~al.]{wainwright2008graphical}
Martin~J Wainwright, Michael~I Jordan, et~al.
\newblock Graphical models, exponential families, and variational inference.
\newblock \emph{Foundations and Trends{\textregistered} in Machine Learning},
  1\penalty0 (1--2):\penalty0 1--305, 2008.

\bibitem[Ye et~al.(2022)Ye, Wu, and Liu]{ye2022first}
Mao Ye, Lemeng Wu, and Qiang Liu.
\newblock First hitting diffusion models.
\newblock \emph{arXiv preprint arXiv:2209.01170}, 2022.

\bibitem[Zhao and Cao(2023)]{zhao2023r}
Zhilin Zhao and Longbing Cao.
\newblock R-divergence for estimating model-oriented distribution discrepancy.
\newblock \emph{arXiv preprint arXiv:2310.01109}, 2023.

\end{thebibliography}

\newpage
\appendix
\onecolumn

\section{List of Detailed Contributions} \label{app:detcontrib}

Main contributions: 
\begin{enumerate}
    \item Firstly, \cite{vargas2023denoising} prompts us to consider Remark \ref{rem:ou}, which allows us to view time reversal in denoising diffusions through the lens of stochastic control. Unfortunately, this did not enable us to directly apply results from \cite{tzen2019theoretical}. The regularity properties for the OU semi-group and the Heat semi-group are significantly different, thus rendering their entropy integral bounds inapplicable.
    
    \item The quantity $\bar{g}_{t,x}(z)$ for the heat semi-group is Lipschitz relative to the Euclidean metric. However, this is not the case for OU. Instead, we can obtain a Lipschitz property with respect to a different metric, results presented in  Section \ref{sec:terminal_cost}:
    \[
    \rho_{OU}\left((t, x),\left(t^{\prime}, x^{\prime}\right)\right)=|| e^{-t} x-x^{\prime} e^{-t^{\prime}} \|+\left|t-t^{\prime}\right|^{1 / 2}
    \]
    
    \item Once we have established a Lipschitz property, the next step is to obtain a covering number for the space of functions induced by $\bar{g}_{t,x}(z)$ (indexed by $x$ and $t$). Unfortunately, as the Lipschitz property is not Euclidean, we can no longer use the simple covering lemma employed in \cite{tzen2019theoretical}:
    \[
    N\left(\mathcal{G}, L^2(Q), \varepsilon\|F\|_{L^2(Q)}\right) \leq N\left(\mathrm{B}^d(R),\|\cdot\|, \varepsilon / 2\right) \cdot N\left([0,1],|\cdot|, \varepsilon^2 / 4\right)
    \]
     This standard result known for covers of function spaces with a Euclidean Lipschitz property \cite{shalev2014understanding} does not apply to our metric space.
    
    \item We then construct the cover for $([0, T] \times B^d(R), \rho_{OU})$ ourselves, which required several non-trivial steps (see the proof for Lemma \ref{lem:covprod}). A significant portion of our work was devoted to this challenging step (the whole process is described in Section \ref{covering_number}):
    \[
    N\left([0, T] \times B^d(R), \rho_{OU}, \epsilon\right) \leq  N\left([0, T],|\cdot|, \epsilon^2 / 4\right) N\left(B^d(R),\|\cdot\|, \epsilon / 2\right)
    \]

    \item Once we obtained step 4, the derived bounds were slightly different, requiring some symbol manipulation and further simplification steps to achieve the final results. In summary, step 4 was a challenging and distinguishing aspect of our work compared to \cite{tzen2019theoretical}. Furthermore, to the best of our knowledge, our work is the first to derive these mathematical properties of the OU semi-group.

    \item Deriving two bounds for OU semi-group in Corollary \ref{lem:first_bound} and Proposition \ref{prop:better_bound}. Since the bound derived in Proposition \ref{prop:better_bound} is tighter we arrived at Observation \ref{obs:tight_bound}, which suggests that for the same parameterization, and target distribution, VP-SDE-based models should perform better than PBM ones.

    \item We derive the transition Kernel for the PBM-SDE which is a generalisation of the F\"{o}llmer drift explored in \citep{tzen2019theoretical} (Section \ref{sec:pbmgm}, Appendix \ref{apdx:PBM} for full derivations), this allowed us to empirically compare PBM-SDE based score matching to VP-SDE in order to see which scales better when limited to the same neural network expressiveness.

    \item Empirically tested Observation \ref{obs:tight_bound} on two datasets of different sizes, as well as on models with varying width dimensions, in Section \ref{sec:simulation}.

\end{enumerate}

\section{Assumptions}\label{assump}

\begin{assumption}\label{assump:a0}
Throughout all this work we assume that the target distribution $\pi$ has a density that is it is absolutely continuous wrt to the Lebesgue measure on $\sR^d$.
\end{assumption}

\begin{assumption}\label{assump:a1}
The function $f$ is differentiable, both $f$ and $\nabla f$ are L-Lipschitz, and there exists a constant $c \in (0,1]$ such that $f \geq c$ everywhere. 
\end{assumption}

\begin{assumption}\label{assump:a2}
The activation function $\sigma : \sR \to \sR$ is differentiable. Moreover,  
 there exists $c_\sigma > 0$ depending only on $\sigma$, such that the following holds: For any L-Lipschitz function $h : \sR \to \R$ which is constant outside the interval $[-R, R]$ and for any $\delta > 0$, there exist real numbers $a$, $\{\alpha_i, \beta_i, \gamma_i\}^m_{i=1}$ where $m \leq c_{\sigma}\frac{RL}{\delta}$, such that the function  $\tilde{h}(x) = a + \sum \alpha_i \sigma(\beta_i x +\gamma_i)$ satisfies $\sup_{x\in \sR}|\tilde{h}(x)-h(x)|\leq \delta$.
\end{assumption}

Finally as per \cite{tzen2019theoretical} we introduce the assumption pertaining to the approximability of $f$ by neural nets. Let $\sigma: \mathbb{R} \rightarrow \mathbb{R}$ be a fixed nonlinearity. Given a vector $w \in \mathbb{R}^n$ and scalars $\alpha, \beta$, define the function
$$
N_{w, \alpha, \beta}^\sigma: \mathbb{R}^n \rightarrow \mathbb{R}, \quad N_{w, \alpha, \beta}^\sigma(x):=\alpha \cdot \sigma\left(w^T x+\beta\right) .
$$
For $\ell \geq 2$, we define the class $\mathcal{N}_{\ell}^\sigma$ of $\ell$-layer feedforward neural nets with activation function $\sigma$ recursively as follows: $\mathcal{N}_2^\sigma$ consists of all functions of the form $x \mapsto \sum_{i=1}^m N_{w_i, \alpha_i, \beta_i}^\sigma(x)$ for all $m \in \mathbb{N}, w_1, \ldots, w_m \in \mathbb{R}^d$, $\alpha_1, \ldots, \alpha_m, \beta_1, \ldots, \beta_m \in \mathbb{R}$, and, for each $\ell \geq 2$,
$$
\begin{aligned}
\mathcal{N}_{\ell+1}^\sigma:= & \bigcup_{k \geq 1} \bigcup_{m \geq 1}\left\{x \mapsto \sum_{i=1}^m N_{w_i, \alpha_i, \beta_i}^\sigma\left(h_1(x), \ldots, h_k(x)\right):\right. \\
& \left.\alpha_1, \ldots, \alpha_m, \beta_1, \ldots, \beta_m \in \mathbb{R}, w_1, \ldots, w_m \in \mathbb{R}^k, h_1, \ldots, h_k \in \mathcal{N}_{\ell}^\sigma\right\} .
\end{aligned}
$$

\begin{assumption}\label{assump:a3}
 For any $R>0$ and $\epsilon >0$, there exist a neural net $\hat{f}\in\mathcal{N}^\sigma _{l,s}$ with $l,s<\text{poly}(1/\epsilon,d,L,R)$, such that
 \begin{align}
     \sup_{x\in B^d(R)}|f(x)-\hat{f}(x)|\leq \epsilon \quad and \quad \sup_{x\in B^d(R)}||\nabla f(x)-\nabla \hat{f}(x)||\leq \epsilon.
 \end{align}
\end{assumption}

\section{Regularity Results}\label{reg}

\begin{repremark}{rem:ou}
The time reversal of the VP-SDE (i.e. $b^{*}(y,t)= -\beta_{T-t} (y -2\sigma^2 \nabla \log \phi_{T-t}(y))$) can be expressed in terms of the OU semigroup via:
\begin{align}
    \nabla \log \phi_{T-t}(y) = \nabla_y \ln U_{T-t}^{\beta_{t}}f(y),
\end{align}
 When $f(x) = \frac{\pi}{\gN(0, \sigma^2 I)}(x)$. This in turn can be related to the score 
 \begin{align}
     \nabla \log p_{T-t}(y) &= -\left(\frac{y}{2\sigma^2}-\nabla \log \phi_{T-t}(y)\right) = -\left(\frac{y}{2\sigma^2}-\nabla_y \ln U_{T-t}^{\beta_{t}}f(y)\right). 
 \end{align}
\end{repremark}
\begin{proof}
Consider the OU semigroup evaluated on the appropriate RND:
\begin{align*}
     U_{t}^{\beta_{t}}f(y) &= \mathbb{E}_{Z \sim \mathcal{N}(0, I)}\left[\frac{\pi}{\gN(0, \sigma^2 I)}\left(e^{- \beta t }y+\sigma(1-e^{-2 \beta t})^{1/2} Z\right)\right] \\
     &= \mathbb{E}_{x_T \sim p^{\mathrm{ref}}_{T|t}(\cdot|x)}\left[\frac{\pi}{\gN(0, \sigma^2 I)}\left(x_T\right)\right] \\ 
     &= \int p^{\mathrm{ref}}_{T|t}(x_T |x) \frac{\pi}{\gN(0, \sigma^2 I)}\left(x_T\right) \mathrm{d}x_T\\
     &= \int \frac{p^{\mathrm{ref}}_{t|T}(x |x_T)  p^{\mathrm{ref}}_T(x_T)}{p^{\mathrm{ref}}_t(x)}\frac{\pi}{\gN(0, \sigma^2 I)}\left(x_T\right) \mathrm{d}x_T \\
     &= \int \frac{p^{\mathrm{ref}}_{t|T}(x |x_T) }{p^{\mathrm{ref}}_t(x)}{\pi}\left(x_T\right) \mathrm{d}x_T = \frac{p_t(x) }{p^{\mathrm{ref}}_t(x)}
\end{align*}
and thus it follows that
\begin{align}
     \nabla \log p_{T-t}(y) = -\left(\frac{y}{2\sigma^2}-\nabla_y \ln U_{T-t}^{\beta_{t}}f(y)\right). 
\end{align}
relating the score and the OU semi-group as required.
\end{proof}

\begin{replemma}{lem:ou_commute}
OU semigroup is commutative with the gradient operator that is for $f: \sR^d \to \sR$ we have $\partial_{y_i } U_t f(y) = U_t \partial_{y_i} f(y)$.
\end{replemma}

\begin{proof}
    It suffices to show that 
\begin{align}
    d(x,z) = \delta^{-1}(f(e^{- t}x+(1-e^{-2 t})^{1/2} z) - f(e^{- t}(x+\delta \ve_i)+(1-e^{-2 t})^{1/2} z)),
\end{align}
is dominated, where $[\ve_i]_j =\delta_{ij}$. As $f$ is Lipchitz by assumption it follows that
\begin{align}
        |d(x,z)| \leq L|\delta^{-1} e^{- t} \delta | =L e^{- t} \leq L
\end{align}
As $L$ is integrable under $\gN(0, I)$ we have shown $d(x,z)$ is dominated for all $\delta$ and thus the partial derivative operator and the OU semigroup commute.
\end{proof}

The choice of $F(z) := L((R \vee 1) + \sqrt{2}||z||)$ with these specific constants arises from the following result.

\begin{replemma}{lem:expected_lip}($\mathscr{L}^2$ Lipchitz condition)
    Let $\bar{g}_{t,x}(z) = g(e^{-t}x + (1-e^{-2t})^{1/2}z) - g(0)$ then it follows that:
    \begin{align*}
        || \bar{g}_{t,x}(z) - \bar{g}_{t',x'}(z)||_{\mathscr{L}^2(Q)} \leq L\left(1 + \sqrt{2}||z||_{\mathscr{L}^2(Q)} \right) \rho_{OU}((t,x), (t',x')) 
    \end{align*}
such that:
\begin{align}
    \rho_{OU}((t,x), (t',x')) = || e^{t}x - x'e^{t'}||  + |t - t'|^{1/2}
\end{align}

\end{replemma}

\begin{proof}
\begin{align*}
 || \bar{g}_{t,x}(z) - \bar{g}_{t',x'}(z)||_{\mathscr{L}^2(Q)} &\leq L || || e^{-t}x + (1-e^{-2t})^{1/2}z - e^{-t'}x' - (1-e^{-2t'})^{1/2}z || ||_{\mathscr{L}^2(Q)}  \\ 
 &\leq L \Big|\Big| || e^{-t}x -e^{-t'}x'||  + |(1-e^{-2t})^{1/2} - (1-e^{-2t'})^{1/2}|\cdot ||z || \Big|\Big|_{\mathscr{L}^2(Q)} \\ 
  &\leq L \Bigg(|| e^{-t}x -e^{-t'}x'||  + |(1-e^{-2t})^{1/2} - (1-e^{-2t'})^{1/2}|\cdot ||z ||_{\mathscr{L}^2(Q)} \Bigg)\\
  &\leq L \Bigg(|| e^{-t}x -e^{-t'}x'||  + |e^{-2t} - e^{-2t'}|^{1/2}\cdot ||z ||_{\mathscr{L}^2(Q)} \Bigg) \\ 
&\leq L \Bigg(|| e^{-t}x -e^{-t'}x'||   + \sqrt{2}|t - t'|^{1/2}\cdot ||z ||_{\mathscr{L}^2(Q)} \Bigg)
 \end{align*}

 Where in the last line we use that $\sup_{t\in [0,T]}|(e^{-2t})'| = 2$ and thus $e^{-2t}$ is 2-Lipchitz.
\end{proof}

\begin{replemma}{lem:envelope}
    Let $g :  R^d \to R$ to L-Lipschitz with respect to the Euclidean norm. Then for $F(z) := L((R \vee 1) + \sqrt{2}||z||)$. 
\begin{align}
    \Big|g\left(e^{- t}x+(1-e^{-2 t})^{1/2} z\right) - g(0)\Big| \leq F(z)
\end{align}
\end{replemma}
\begin{proof}
By Lipschitz continuity for all $z \in R^d, X\in B^d(R), t\in [0,T]$ we have:

\begin{align}
    |g\left(e^{- t}x+(1-e^{-2 t})^{1/2} z\right) - g(0)| &\leq L || e^{- t}x+(1-e^{-2 t})^{1/2} z|| \\
    &\leq L ( e^{- t}||x||+(1-e^{-2 t})^{1/2} ||z||)
\end{align}  

Since both $e^{- t}$ and $(1-e^{-2 t})^{1/2}$ are strictly smaller than $1$, we have:
\begin{align}
L ( e^{- t}||x||+(1-e^{-2 t})^{1/2} ||z||) &\leq L(R + ||z||) \\
&\leq L((R\vee 1) + ||z||) \leq F(z)
\end{align}  
\end{proof}

\section{Covering Number Results}\label{covering}

\begin{remark}\label{rem:metricspace}
    The space $([0,T]\times B^d(R),\rho_{OU})$ is a metric space, where 
\begin{align}
    \rho_{OU}((t,x), (t',x')) = || e^{-t}x - x'e^{-t'}||  + |t - t'|^{1/2}.
\end{align}
\end{remark}

\begin{proof}

\begin{itemize}
    \item \textbf{Positive definiteness}:
\begin{align}
    \rho_{OU}((t,x), (t',x')) &= 0 \Longleftrightarrow \\
    \label{eq::base}
    ||e^{-t}x-x'e^{-t'}|| + |t-t'|^{1/2} &= 0 \Longleftrightarrow \\
    \label{eq::conc}
    x=x' \text{ and } t&=t.
\end{align}
Since in (\ref{eq::base}) both terms are positive on the LHS, each has to be $0$ to get the RHS, thus we get (\ref{eq::conc}). 

\item \textbf{Symmetry}:
\begin{align}
    \rho_{OU}((t,x),(t',x'))=\rho_{OU}((t',x'),(t,x)).
\end{align}

\item \textbf{Triangle inequality}:
we show triangle inequality on $(t,x),(t',x')$ and $(t'',x'')$. First let us note, that $||e^{-t}x-x'e^{-t'}||+||e^{-t'}x'-x''e^{-t''}||\geq ||e^{-t}x-x''e^{-t''}||$, since $||\cdot||$ has the triangle inequality. Now:
\begin{align}
    |t-t'|^{1/2}+|t'-t''|^{1/2} &\geq |t-t''|^{1/2} \Longleftrightarrow \\
    \label{eq::end}
    |t-t'|+2|t-t'|^{1/2}|t'-t''|^{1/2}+|t'-t''|&\geq |t-t''|.
\end{align}
(\ref{eq::end}) is true, since $|\cdot|$ has the triangle inequality and $2|t-t'|^{1/2}|t'-t''|^{1/2}\geq 0$.
\end{itemize}
\end{proof}

\begin{replemma}{lem:metlip}
Given the metric space $\big( [0,T] \times B^d(R) , \rho_{OU}\big)$ where:

\begin{align}
    \rho_{OU}((t,x), (t',x')) = || e^{-t}x - x'e^{-t'}||  + |t - t'|^{1/2}
\end{align}
and
\begin{align}
    ||(t,x)||_{OU} =  \rho_{OU}((t,x), (0, 0))= || e^{-t}x ||  + |t|^{1/2}
\end{align}
It follows that:
\begin{align}
     N(\gG,  \mathscr{L}^2(Q), \epsilon ||F ||_{\mathscr{L}^2(Q)}) \leq  N([0,T] \times B^d(R),  \rho_{OU}, \epsilon) 
\end{align}

\end{replemma}
\begin{proof}

Consider the $\epsilon$-cover $A_{\rho_{OU}}$ with respect to $\rho_{OU}$ of $[0,T] \times B^d(R)$ it follows that for any $(t,x) \in [0,T] \times B^d(R)$ we have that there exists $(t',x') \in A_{\rho_{OU}}$  such that $\rho_{OU}((t,x), (t',x'))  \leq \epsilon$ then by Lemma \ref{lem:expected_lip} it follows that 
\begin{align}
 || \bar{g}_{t,x}(z) - \bar{g}_{t',x'}(z)||_{\mathscr{L}^2(Q)} &\leq L\left(1 + \sqrt{2}||z||_{\mathscr{L}^2(Q)} \right) \rho_{OU}((t,x), (t',x'))  \\ 
& \leq ||F ||_{\mathscr{L}^2(Q)} \rho_{OU}((t,x), (t',x'))  \\
& \leq ||F ||_{\mathscr{L}^2(Q)} \epsilon 
\end{align}
Hence the set:
\begin{align}
    \gG_{\rho_{OU}} = \{ \bar{g}_{t,x}:(t,x) \in A_{\rho_{OU}}\}
\end{align}
is an $||F || \epsilon$ cover of $\gG$ with respect to the metric $\rho_{OU}$
\end{proof}

\begin{replemma}{lem:covprod}
We have that
\begin{align}
     N([0,T] \times B^d(R),  \rho_{OU}, \epsilon)  \leq  N([0,T], |\cdot|,  \epsilon^2/4) N(B^d(R), ||\cdot ||, \epsilon/2)
\end{align}
\end{replemma}
\begin{proof}

Let $B^d_{r_0}(R)$ denote a euclidean d-dimensional ball of radius $R$ centered at $r_0$ and let $B^{d+1}_{t_0 \oplus x_0, \rho}(R')$\footnote{$a\oplus b$ denotes the concatenation of $a$ and $b$.} denote it's counterpart with respect to the metric $\rho$.  Now notice that if $|| e^{-t} x- e^{-t_0}x_0||  + | t-t_0|^{1/2} \leq \epsilon$ then $|| e^{-t_0}  (x-x_0)||  \leq \epsilon$ and  $|| x-x_0||  \leq  e^{t_0} \epsilon,$ thus,
\begin{align}
   \{t_0\}  \times B^d_{x_0}(\epsilon) \subseteq \{t_0\}  \times  B^d_{x_0}(e^{t_0} \epsilon) \subseteq  B^{d+1}_{t_0 \oplus x_0, \rho}(\epsilon),
\end{align}
then since $\{t_0\}  \times B^d_{x_0}(e^{t_0} \epsilon) \subseteq  B^{d+1}_{t_0 \oplus x_0, \rho}(\epsilon)$ we can construct an $\epsilon$ cover namely $A_{t_0}$ of $\{t_0\} \times B^d(R)$ with $N(B^d(R), ||\cdot ||, \epsilon e^{t_0})$ balls. Finally notice that if $|| e^{-t}  x- e^{-t_0}x_0||  + | t-t_0|^{1/2} \leq \epsilon$ it follows that $|t-t_0|^{1/2} \leq \epsilon$ thus $[0,T]$ can be covered in  $N([0,T], |\cdot|,  \epsilon^2) \leq T\epsilon^{-2}$ sub intervals.

Let $U_T$ be the smallest cover containing $N([0,T], |\cdot|,  \epsilon^2)$ intervals $u_n$ each centered at $t_n$ , then: 
\begin{align}
   A =  \bigcup_{u_n \in U_T} A_{t_n} 
\end{align} 

is an $\epsilon$ cover of $[0, T] \times B^d(R)$ (with respect to the metric $\rho_{OU}$), notice this follows as $\forall x \in B^d(R)$ there exists an $x_0$ such that 
\begin{align}
[t_n -\epsilon^2, t_n+ \epsilon^2] \times \{x\} \subseteq B^{d+1}_{t_n\oplus x_0, \rho}(\epsilon) \in  A_{t_n}
\end{align}

Now we can see that 
\begin{align}
|A| \leq  |U_T| |A_0| &= N([0,T], |\cdot|,  \epsilon^2) N(B^d(R), ||\cdot ||, \epsilon e^{t_0}), \\
&\leq  N([0,T], |\cdot|,  \epsilon^2/4) N(B^d(R), ||\cdot ||, \epsilon/2) ,
\end{align}
where $|A_0 |=\max_{n}|A_{t_n} |$, completing our proof. \\

\end{proof}

\begin{replemma}{lem:first_bound}
        Given the metric space $\big( [0,T] \times B^d(R) , \rho_{OU}\big)$ it follows that:
    \begin{align}
         N(B^d(R), ||\cdot ||, \epsilon/2) N([0,T], |\cdot|,  \epsilon^2/4)\leq \left(\frac{2\sqrt{3R T}}{\epsilon}\right)^{2d}
    \end{align}
\end{replemma}

\begin{proof}
We know the covering number of $ N(B^d(R), ||\cdot ||, \epsilon)$ is $\left(\frac{2R}{\epsilon}\right)^d$, and for $N([0,T], |\cdot|, \epsilon)$, it is $\frac{T}{2\epsilon}$. In our settings:

\begin{align}
    N(B^d(R), ||\cdot ||, \epsilon/2) N([0,T], |\cdot|,  \epsilon^2/4)\leq 
    \left(\frac{6R}{\epsilon}\right)^d \left(\frac{2T}{\epsilon^2}\right)
\end{align}

Now for $d \geq 2$ and $\epsilon$ small enough ($\epsilon^{d-2} \leq (2T)^{d-1} $)  we get $\left(\frac{2T}{\epsilon^2}\right) \leq \left(\frac{2T}{\epsilon}\right)^d  $
After this, our inequality will become:

$$\left(\frac{6R}{\epsilon}\right)^d \left(\frac{2T}{\epsilon^2}\right) \leq \left(\frac{6R}{\epsilon}\right)^d \left(\frac{2T}{\epsilon}\right)^d 
 = \left(\frac{2\sqrt{3R T}}{\epsilon}\right)^{2d}$$
 
\end{proof}

\begin{repproposition}{prop:better_bound}
    Given the metric space $\big( [0,T] \times B^d(R) , \rho_{OU}\big)$ it follows that:
    \begin{align}
         N([0,T] \times B^d(R),  \rho_{OU}, \epsilon)  \leq \left( \frac{ 2   e^{-\epsilon^2/2}  \sqrt{3TR}}{\epsilon}\right)^{d}
    \end{align}
\end{repproposition}
\begin{proof}

Let $\rho_{OU}=\rho$ and $B^d_{r_0}(R)$ denote a euclidean d-dimensional ball of radius $R$ centered at $r_0$ and let $B^{d+1}_{t_0 \oplus x_0, \rho}(R')$ denote it's counterpart with respect to the metric $\rho$.  Now notice that if $|| e^{-t} x- e^{-t_0}x_0||  + | t-t_0|^{1/2} \leq \epsilon$ then $|| e^{-t_0}  (x-x_0)||  \leq \epsilon$ and  $|| x-x_0||  \leq  e^{t_0} \epsilon,$ thus,
\begin{align}
   \{t_0\}  \times B^d_{x_0}(\epsilon) \subseteq \{t_0\}  \times  B^d_{x_0}(e^{t_0} \epsilon) \subseteq  B^{d+1}_{t_0 \oplus x_0, \rho}(\epsilon),
\end{align}
then since $\{t_0\}  \times B^d_{x_0}(e^t \epsilon) \subseteq  B^{d+1}_{t \oplus x_0, \rho}(\epsilon)$ we can construct an $\epsilon$ cover namely $A_{t_0}$ of $\{t_0\} \times B^d(R)$ with $\left(  6 R{\epsilon }^{-1} e^{-t_0} \right)^{d}$ balls of form the form $B^{d+1}_{t_0 \oplus x_0, \rho}$. Finally notice that if $|| e^{-t}  x- e^{-t_0}x_0||  + | t-t_0|^{1/2} \leq \epsilon$ it follows that $|t-t_0|^{1/2} \leq \epsilon$ thus $[0,T]$ can be covered in $2^{-1}T \epsilon^{-2}$. 


picking the cover $U_T$ such that its elements $u_n$ are centered at $(n + 1) \epsilon^2/2$ , then: 
\begin{align}
   A =  \bigcup_{u_n \in U_T} A_{(n + 1) \epsilon^2/2} 
\end{align}
is an $\epsilon$ cover of $[0, T] \times B^d(R)$ (with respect to the metric $\rho_{OU}$), notice this follows as $\forall x \in B^d(R)$ there exists an $x_0$ such that 
\begin{align}
[(n + 1) \epsilon^2/2 -\epsilon^2, (n + 1) \epsilon^2/2 + \epsilon^2] \times \{x\} \subseteq B^{d+1}_{(n + 1) \epsilon^2/2 \oplus x_0, \rho}(\epsilon), 
\end{align}
with  $B^{d+1}_{(n + 1) \epsilon^2/2 \oplus x_0, \rho} \in A_{(n + 1) \epsilon^2/2}$.

Now we can see that $|A| \leq  |U_T| |A_0|$ ( $|A_0 |=\max_{n}|A_{(n + 1) \epsilon^2/2} |$) completing our proof.
\end{proof}

From Lemmas \ref{lem:metlip}, \ref{lem:covprod} it follows that :
\begin{align}
      N(\gG,  \mathscr{L}^2(Q), \epsilon ||F ||_{\mathscr{L}^2(Q)}) \leq N([0,T], |\cdot|,  \epsilon^2/4) N(B^d(R), ||\cdot ||, \epsilon/2)
\end{align}

\begin{lemma}
    The Koltchinskii-Pollard $\varepsilon$-entropy of $N(\gG,  \mathscr{L}^2(Q), \epsilon ||F ||_{\mathscr{L}^2(Q)}) $ is given by
$$
H(\mathcal{G}, F, \varepsilon):=\sup _Q \sqrt{\log 2 N\left(\mathcal{G}, L^2(Q), \varepsilon\|F\|_{L^2(Q)}\right)}
$$
Then we have 
$$
J(\gG,  \mathscr{L}^2(Q))=\int_0^{\infty}  H(\mathcal{G}, F, \varepsilon) \mathrm{d} \varepsilon \leq 2 \sqrt{3 \pi R d T} 
$$ with $H(\mathcal{G}, F, \varepsilon) \leq \sqrt{ \left(4 d \log \frac{2 \sqrt{3 R T}}{\varepsilon}\right)_{+}}$. 
\end{lemma}
\begin{proof}
 Following the derivations from \cite{tzen2019theoretical}, and our bound from Lemma \ref{lem:first_bound}:

\begin{align}
    J(\gG,  \mathscr{L}^2(Q))=\int_0^{\infty}  H(\mathcal{G}, F, \varepsilon) \mathrm{d} \varepsilon \leq  \int_0^{\infty} \sqrt{ \left(4 d \log \frac{2 \sqrt{3 R T}}{\varepsilon}\right)_{+}} d\epsilon 
\end{align}

\begin{align}
    = 2 \sqrt{d}  \int_0^{2\sqrt{3R}} \sqrt{ \left(\log \frac{2 \sqrt{3 R T}}{\varepsilon}\right)} d\epsilon = 
\end{align}

\begin{align}
    =4 \sqrt{3dR T} (ye^{-y^2} \Big{|}^{0}_{\infty} - \int^{0}_{\infty} e^{-y^2})= 4 \sqrt{3dR T} \frac{\sqrt{\pi}}{2} = 2\sqrt{3dR\pi T}
\end{align}

\end{proof}

Thus by Lemma D.4  (see the start of Page 18 in \cite{tzen2019neural}) we now have the following corollary of Lemma C.4 from \cite{tzen2019theoretical}
\begin{corollary}\label{col:tzenc4}(Theorem C.4. from \cite{tzen2019theoretical}) Let $g: \mathbb{R}^d \rightarrow \mathbb{R}$ be L-Lipschitz with respect to the Euclidean norm. Let $Z_1, \ldots, Z_N$ be i.i.d. copies of a d-dimensional random vector $Z$, such that $U:=\|Z\|$ has finite $\psi_2$ norm. Then there exists an absolute constant $C>0$, such that, for any $\gamma>0$,
$$
\begin{aligned}
& \sup _{x \in \mathrm{B}^d(R)} \sup _{t \in[0,1]}\left|\frac{1}{\textcolor{magenta}{N}} \sum_{n=1}^{\textcolor{magenta}{N}} \textcolor{magenta}{g\left(e^{- t}x+(1-e^{-2 t})^{1/2} Z_n\right)}-\mathbb{E}\left[\textcolor{magenta}{g\left(e^{- t}x+(1-e^{-2 t})^{1/2} Z\right)}\right]\right| \\
& \leq C\left[\frac{16 L \sqrt{6 \pi R d}\left((R \vee 1)+\|U\|_{\psi_2}\right)}{\sqrt{\textcolor{magenta}{N}}}+5 L\left((R \vee 1)+\|U\|_{\psi_2}\right) \sqrt{\frac{\gamma}{\textcolor{magenta}{N}}}\right]
\end{aligned}
$$
with probability at least $1-e^{-\gamma}$.
\end{corollary}

Finally Theorem C.1 in \cite{tzen2019theoretical} will hold true in our setting, with the modified choice of 
\begin{align}
 N=\left\lceil\left(\frac{C \sqrt{d}}{\varepsilon} \cdot L((R \vee 1)+ { \color{magenta}\sqrt{2d}}+\sqrt{6}) \cdot(16 \sqrt{6 \pi R d \color{magenta}T}+5 \sqrt{\log 4(d+1)})\right)^2\right\rceil,
\end{align}
For completenteness we will now restate our adaptation of Theorem C.1.

\begin{repcorollary}{col:tzen}(Theorem C.1. from \cite{tzen2019theoretical})
    
 For any $\varepsilon>0$ and any $R>0$, there exist ${\color{magenta}N}=\operatorname{poly}(1 / \varepsilon, d, L, R, \color{magenta}T )$ points $z_1, \ldots, z_{{\color{magenta}N}} \in \mathbb{R}^d$, for which the following holds:
\begin{align*}
\max _{n \leq N}\left\|z_n\right\| \leq 8 \sqrt{(d+6) \log {\color{magenta}N}} \\
\sup _{x \in \mathrm{B}^d(R)} \sup _{t \in[0,1]}\left|\frac{1}{{\color{magenta}N}} \sum_{n=1}^{{\color{magenta}N}} \textcolor{magenta}{f\left(e^{- t}x+(1-e^{-2 t})^{1/2} z_n\right)}-U_t f(x)\right| \leq \varepsilon \\
\sup _{x \in \mathrm{B}^d(R)} \sup _{t \in[0,1]}\left\|\frac{1}{{\color{magenta}N}} \sum_{n=1}^{{\color{magenta}N}} \nabla \textcolor{magenta}{f\left(e^{- t}x+(1-e^{-2 t})^{1/2} z_n\right)}-\nabla U_t f(x)\right\| \leq \varepsilon
\end{align*}
\end{repcorollary}

We now have everything that is required to show the neural network approximation results. 
\begin{remark}\label{rem:bounding_remark}
    The same computation for our tight-bound from Proposition \ref{prop:better_bound} leads to:
\begin{align}
   H(\mathcal{G}, F, \varepsilon) \leq \sqrt{2d\ln\left( \frac{e^{-\epsilon^2/2}  \sqrt{3TR}}{\epsilon}\right)_{+}} 
\end{align}
Moving forward:
 $$ J(\mathcal{G}, F) \leq \int_{0}^{\sqrt{W(1)}} \sqrt{2d\ln\left( \frac{e^{-\epsilon^2/2}  \sqrt{3TR}}{\epsilon}\right)_{+}}  d\epsilon $$
 Where $W(1)$ is the solution to $-x=\ln x$. Unfortunately, we weren't able to find a closed-form solution to this integral.

\end{remark}





\section{Neural Network Approximation}\label{apdx:approx}

\begin{corollary}\label{reg:corr}
Under Assumption \ref{assump:a1}, the vector field $\nabla \log U_t f(x)$ is bounded in norm by $\frac{L}{c}$ and is Lipschitz with  constant $\frac{L}{c}+ \frac{L^2}{c^2}$ where L is the max of the Lip constant of $f$ and $\nabla f$.
\end{corollary}
\begin{proof}
    By direct application of Lemma B.1. (\cite{tzen2019theoretical}) and our Lemma \ref{lem:ou_commute}, which assures that OU semi-group commutes with the gradient operator, we have that the results of this Corollary hold. 
\end{proof}

We now proceed to adapt one of the main theorems in \cite{tzen2019theoretical}. Whilst the changes are minor to the sketch in \cite{tzen2019neural} some are subtle thus we have incorporated this proof for completeness. We highlight in {\color{magenta} magenta} the subtle changes required to adapt the result.

\begin{corollary}(Tzen and Ragisnky)
\label{cor:th3.2}
 Let  $0<\varepsilon<4 L / c$   and  $R>0$ be given. Then there exists a neural net $\widehat{v}: \mathbb{R}^d \times[0,1] \rightarrow \mathbb{R}^d$ of size polynomial in $1 / \varepsilon, d, L, R, c, 1 / c$, such that the activation function of each neuron is an element of the set $\left\{\sigma, \sigma^{\prime}, \operatorname{ReLU}\right\}$, and the following holds:
$$
\sup _{x \in \mathrm{B}^d(R)} \sup _{t \in[0,1]}\left\|\widehat{v}(x, {t})-\nabla \log U_t f(x)\right\| \leq \varepsilon
$$
and
$$
\max _{i \in[d]} \sup _{x \in \mathbb{R}^d} \sup _{t \in[0,1]}\left|\widehat{v}_i(x, {t})\right| \leq \frac{2 L}{c} .
$$
\end{corollary} 
\begin{proof}
Let $\delta=\frac{c^2 \varepsilon}{16 L}$. By Theorem C.1 (which has been proved to hold true in our settings in Appendix C), there exist points $z_1, \ldots, z_N \in \mathbb{R}^d$ with $N=\operatorname{poly}(1 / \delta, d, L, R)$, such that $R_{N, d}:=\max _{n \leq N}\left\|z_n\right\| \leq 8 \sqrt{(d+6) \log N}$, and the function $\varphi: \mathbb{R}^d \times[0,1] \rightarrow \mathbb{R}$ defined by
\begin{align}
{\color{magenta}\varphi(x, t)\coloneqq \frac{1}{N} \sum_{n=1}^N f\left(e^{-t}x+ (1- e^{-2t})^{1/2} z_n\right)}
\end{align}
satisfies
$$
\sup _{x \in \mathrm{B}^d(R)} \sup _{t \in[0,1]}\left|\varphi(x, t)-U_t f(x)\right| \leq \delta \quad \text { and } \quad \sup _{x \in \mathrm{B}^d(R)} \sup _{t \in[0,1]}\left\|\nabla \varphi(x, t)-\nabla U_t f(x)\right\| \leq \delta
$$

By Assumption \ref{assump:a3}, there exists a neural net $\widehat{f}: \mathbb{R}^d \rightarrow \mathbb{R}$ be that approximates $f$ and the gradient of $f$ to accuracy $\delta$ on the blown-up ball $\mathrm{B}^d\left(R+R_{N, d}\right)$. Then the function
$$
\widehat{\varphi}: \mathbb{R}^d \times[0,1] \rightarrow \mathbb{R}, \quad  {\color{magenta}\widehat{\varphi}(x, t):=\frac{1}{N} \sum_{n=1}^N \widehat{f}\left(e^{-t}x+ (1- e^{-2t})^{1/2} z_n\right)}
$$
can be computed by a neural net of $\operatorname{size} N \cdot \operatorname{poly}(1 / \delta, d, L, R)$, such that

\begin{align}
\begin{aligned}
& \sup _{x \in \mathrm{B}^d(R)} \sup _{t \in[0,1]}\left|\widehat{\varphi}(x, t)-U_t f(x)\right| \\
& \leq \sup _{x \in \mathrm{B}^d(R)} \sup _{t \in[0,1]}|\widehat{\varphi}(x, t)-\varphi(x, t)|+\sup _{x \in \mathrm{B}^d(R)} \sup _{t \in[0,1]}\left|\varphi(x, t)-U_t f(x)\right| \\
&\leq {\color{magenta}\sup _{x \in \mathrm{B}^d(R)} \sup _{t \in[0,1]} } {\color{magenta}\left|\frac{1}{N} \sum_{n=1}^N \widehat{f}\left(x+ (1- e^{-2t})^{1/2} z_n\right)-\frac{1}{N} \sum_{n=1}^N {f}\left(x+ (1- e^{-2t})^{1/2} z_n\right)\right|}\nonumber\\
 &\quad \quad\quad +\sup _{x \in \mathrm{B}^d(R)} \sup _{t \in[0,1]}\left|\varphi(x, t)-U_t f(x)\right| \\
& \quad \leq \sup _{x \in \mathrm{B}^d\left(R+R_{N, d}\right)}|\widehat{f}(x)-f(x)|+\sup _{x \in \mathrm{B}^d(R)} \sup _{t \in[0,1]}\left|\varphi(x, t)-U_t f(x)\right| \leq 2 \delta
\end{aligned}
\end{align}

where the third inequality follows since $ {\color{magenta}e^{-t} \in [0,1]}$ and the final inequality follows since
\begin{align*}
 {\color{magenta}\max_n\sup_{t \in [0,1]} (1- e^{-2t})^{1/2} ||z_n|| =\max_n  ||z_n|| = R_{N,d}}
\end{align*}
Similarly
\begin{align*}
\begin{aligned}
& \sup _{x \in \mathrm{B}^d(R)} \sup _{t \in[0,1]}\left\|\nabla \widehat{\varphi}(x, t)-\nabla U_t f(x)\right\| \\
& \leq \sup _{x \in \mathrm{B}^d(R)} \sup _{t \in[0,1]}\|\nabla \widehat{\varphi}(x, t)-\nabla \varphi(x, t)\|+\sup _{x \in \mathrm{B}^d(R)} \sup _{t \in[0,1]}\left\|\nabla \varphi(x, t)-\nabla U_t f(x)\right\| \\
& \quad \leq \sup _{x \in \mathrm{B}^d\left(R+R_{N, d}\right)}\|\nabla \widehat{f}(x)-\nabla f(x)\|+\sup _{x \in \mathrm{B}^d(R)} \sup _{t \in[0,1]}\left\|\nabla \varphi(x, t)-\nabla U_t f(x)\right\| \leq 2 \delta .
\end{aligned}
\end{align*}

Since $f$ is $L$-Lipschitz and bounded below by $c$, we have $U_t f(x) \geq \E_{Z \sim \mathcal{N}(0, I)}[c] = c$, and
\begin{align*}
    {\color{magenta}U_t f(x) =  \E_{Z \sim \mathcal{N}(0, I)} \left[f( e^{-t}x +(1-e^{-2t} )^{1/2} Z )\right] }&{\color{magenta}\leq \E_{Z \sim \mathcal{N}(0, I)} \left[ L(||x|| + \sqrt{2} ||z||) + f(0)\right]} \\
     &{\color{magenta}= L||x|| + f(0) + L \sqrt{2}  \E[||z||]} \\
    &{\color{magenta}\leq L(||x|| +  \sqrt{2d}) + f(0)}
\end{align*}
Thus it follows that
${\color{magenta}c \leq U_t f(x) \leq L(\|x\|+\sqrt{2d})+f(0)}$ for any $x \in \mathbb{R}^d$ and $t \in[0,1]$. Therefore, on $\mathrm{B}^d(R) \times[0,1]$,
$$
{\color{magenta}\frac{c}{2} \leq \widehat{\varphi}(x, t) \leq L(R+\sqrt{2d})+f(0)+\frac{c}{2}}
$$

where we use $\delta \leq c/4$. Without loss of generality, we may assume that $L \geq 1$. Then, for any $x \in \mathrm{B}^d(R)$ and $t \in[0,1]$
$$
\begin{aligned}
& \left\|\nabla \log \widehat{\varphi}(x, t)-\nabla \log U_t f(x)\right\| \\
& =\left\|\frac{\nabla \widehat{\varphi}(x, t)}{\widehat{\varphi}(x, t)}-\frac{\nabla U_t f(x)}{U_t f(x)}\right\| \\
& \leq \frac{1}{\widehat{\varphi}(x, t)}\left\|\nabla \widehat{\varphi}(x, t)-\nabla U_t f(x)\right\|+\left\|\frac{\nabla U_t f(x)}{U_t f(x)}\right\| \frac{\left|\widehat{\varphi}(x, t)-U_t f(x)\right|}{\widehat{\varphi}(x, t)} \\
& \leq \frac{2 L}{c} \cdot 2 \delta+\frac{L}{c} \cdot \frac{2}{c} \cdot 2 \delta \\
& \leq \frac{\varepsilon}{2},
\end{aligned}
$$
where we have used Corollary \ref{reg:corr} to bound $\left\|\frac{\nabla U_t f}{U_t f}\right\| \leq L / c$. In other words, $\nabla \log \widehat{\varphi}(x, t)$ approximates $\nabla \log U_t f(x)$ to accuracy $\varepsilon / 2$ uniformly on $\mathrm{B}^d(R) \times[0,1]$. It remains to approximate $\nabla \log \widehat{\varphi}(x, t)$ by a neural net to accuracy $\varepsilon / 2$.

To that end, we first represent $\nabla \log \widehat{\varphi}(x, t)$ as a composition of several elementary operations and then approximate each step by a neural net. Specifically, the computation of $v_i=\partial_i \log \widehat{\varphi}(x, t)$ can be represented as a computation graph with the following structure:
\begin{enumerate}
    \item  Compute $a=\widehat{\varphi}(x, t)$.
    \item Compute $b_i=\partial_i \widehat{\varphi}(x, t)$.
    \item Compute $r=1 / a$.
    \item  Compute $v_i=r b_i$.
\end{enumerate}
Given $x$ and $t, a$ is computed by a neural net with activation function $\sigma$, of size $\operatorname{poly}(1 / \delta, d, L, R)$ and depth poly $(1 / \delta, d, L, R)$. Therefore, by the cheap gradient principle (Lemma D.1 from \cite{tzen2019theoretical}), $b_i$ can be computed by a neural net of size poly $(1 / \delta, d, L, R)$, where the activation function of each neuron is an element of the set $\left\{\sigma, \sigma^{\prime}\right\}$. Next, since $a$ takes values in $[c / 2, L(R+\sqrt{2d})+f(0)+c / 2]$, by Lemma D.2 from \cite{tzen2019theoretical} the reciprocal $r=1 / a$ can be computed to accuracy $\varepsilon /(4 L \sqrt{d})$ by a 2 -layer neural net with activation function $\sigma$ and of size
$$
\mathcal{O}\left(\frac{4}{c^2} \cdot\textcolor{magenta}{(L(R+\sqrt{2d})+f(0)+c / 2) }\cdot \frac{4 L \sqrt{d}}{\varepsilon}\right) \leq \operatorname{poly}(1 / \varepsilon, d, L, R, c, 1 / c)
$$
Let $\widehat{r}$ denote the resulting approximation. Then, since $\left|b_i\right| \leq 2 L$ and $|\widehat{r}| \leq 2 / c+\varepsilon /(4 L \sqrt{d}) \leq 4 / c$, by Lemma D.2 the product $\widehat{r} b_i$ can be approximated to accuracy $\varepsilon / 4 \sqrt{d}$ by a 2-layer neural net with activation function $\sigma$ and with at most
$$
\mathcal{O}\left((4 / c \vee 2 L)^2 \cdot \frac{4 \sqrt{d}}{\varepsilon}\right) \leq \operatorname{poly}(1 / \varepsilon, d, L, 1 / c)
$$
neurons. The overall accuracy of the approximation is
$$
\left|\widehat{v}_i-v_i\right| \leq\left|\widehat{v}_i-\widehat{r} b_i\right|+\left|\widehat{r} b_i-r b_i\right| \leq \frac{\varepsilon}{2 \sqrt{d}}
$$
Thus, the vector $v=\left(v_1, \ldots, v_d\right)$ can be $\varepsilon / 2$-approximated by $\tilde{v}(x, t)$, where $\tilde{v}: \mathbb{R}^d \times[0,1] \rightarrow \mathbb{R}^d$ is a neural net with vector-valued output that has the $\operatorname{size} \operatorname{poly}(1 / \varepsilon, d, L, R, c, 1 / c)$. Finally, since $\sup _{x \in \mathrm{B}^d(R)} \sup _{t \in[0,1]}\left|\tilde{v}_i(x, t)\right| \leq 2 L / c$, the function
$$
\widehat{v}_i(x, t):=\min \left\{\max \left\{\tilde{v}_i(x, t),-2 L / c\right\}, 2 L / c\right\}
$$
is continuous, takes values in $[-2 L / c, 2 L / c]$ and coincides with $\tilde{v}_i$ on $\mathrm{B}^d(R) \times[0,1]$. Moreover, the min and max operations can each be implemented exactly using $\mathcal{O}(1)$ ReLU neurons.
\end{proof}

\begin{repcorollary}{col:est}
Suppose Assumptions 1-3 are in force. Let L denote the maximum of the Lipschitz constants of $f$ and $\nabla f$. Then for all $0< \epsilon < 16L^2/c^2$, there exists a neural net $\hat{v} : R^d \times [0,1] \to R^d$ with size polynomial in $1/\epsilon, d, L, c, 1/c$ such that the activation function  of each neuron in the set of $\{\sigma, \sigma', ReLU\}$, and the following hold: If $\{\hat{x_t}\}_{t\in[0,1]}$ is the diffusion process governed by the It\^o SDE:
\begin{align} 
d\hat{x}_t = \hat{b}(\hat{x}_{t}, t)\dd t + \sqrt{2 \beta} \dd W_t
\end{align}
with $x_0 \sim p_1 \approx \gN(0, I)$ with the drift $\hat{b}(x,t) = - (x - 2 \hat{v}(x, 1-t))$, then $\hat{\mu} := \mathrm{Law}(\hat{x}_1)$, satisfies $D(\mu||\hat{\mu}) \leq \epsilon$.
\end{repcorollary}
\begin{proof}

 For any $R>0$, Corollary \ref{cor:th3.2} guarantees the existence of a neural net $\widehat{v}: \R^d \times[0,1] \rightarrow \R^d$ that satisfies
\begin{equation}\label{eq:error_bound}
    \sup _{x \in \mathrm{B}^d(R)} \sup _{t \in[0,1]}\left\|\widehat{v}(x, {t})-\nabla \log U_t f(x)\right\| \leq \sqrt{\varepsilon}
\end{equation}
 and
\begin{align}\label{eq:network_bound}
     \max _{i \in[d]} \sup _{x \in \mathbb{R}^d} \sup _{t \in[0,1]}\left|\widehat{v}_i(x, {t})\right| \leq \frac{2 L}{c} .
\end{align}

 Let $\boldsymbol{\mu}:=\operatorname{Law}\left(x_{[0,1]}\right)$ and $\widehat{\boldsymbol{\mu}}:=\operatorname{Law}\left(\widehat{x}_{[0,1]}\right)$. The Girsanov formula gives
$$
\KL(\boldsymbol{\mu} \| \widehat{\boldsymbol{\mu}})=\frac{1}{2} \int_0^1 \mathbf{E}\left\|b\left(x_t, t\right)-\widehat{b}\left(x_t, t\right)\right\|^2 \mathrm{~d} t
$$

where the interchange of the integral and the expectation follows from Fubini's theorem because both $b$ and $\widehat{b}$ are bounded by Corollary \ref{reg:corr} and (\ref{eq:network_bound}). We now proceed to estimate the integrand. For each $t \in[0,1]$
$$
\begin{aligned}
& \mathbf{E}\left\|b\left(x_t, t\right)-\widehat{b}\left(x_t, t\right)\right\|^2 \\
& =\mathbf{E}\left[\left\|b\left(x_t, t\right)-\widehat{b}\left(x_t, t\right)\right\|^2 \cdot \mathbf{1}\left\{x_t \in \mathrm{B}^d(R)\right\}\right]+\mathbf{E}\left[\left\|b\left(x_t, t\right)-\widehat{b}\left(x_t, t\right)\right\|^2 \cdot \mathbf{1}\left\{x_t \notin \mathrm{B}^d(R)\right\}\right] \\
& =: T_1+T_2,
\end{aligned}
$$
where $T_1 \leq \varepsilon$ by (\ref{eq:network_bound}). To estimate $T_2$, we first observe that, since the OU drift is bounded in norm by $L / c$ by \ref{reg:corr}, we have
$$
\mathbf{P}\left\{\sup _{t \in[0,1]}\left\|x_t\right\| \geq R\right\} \leq \frac{\sqrt{d}+L / c}{R}
$$
(\cite{bubeck2018sampling}, Lemma 3.8). Therefore,
$$
T_2 \leq \frac{9 d L^2}{c^2} \cdot \frac{\sqrt{d}+L / c}{R}
$$
Since some of the bounds differ from the original \cite{tzen2019theoretical} we verify that the bound still holds for our drift. We used that $d \geq 2$.
\begin{align*}
    \begin{aligned}
        T_2 = \mathbf{E}\left[\left\|b\left(x_t, t\right)-\widehat{b}\left(x_t, t\right)\right\|^2 \cdot \mathbf{1}\left\{x_t \notin \mathrm{B}^d(R)\right\}\right] = \int_{x_t \notin \mathrm{B}^d(R)} \|b\left(x_t, t\right)-\widehat{b}\left(x_t, t\right)\|^2 dP_{x_t} =
        \\= \int_{x_t \notin \mathrm{B}^d(R)}
        2\|b\left(x_t, t\right)\|^2+2\|\widehat{b}\left(x_t, t\right)\|^2 dP_{x_t} 
        \leq 
        \int_{x_t \notin \mathrm{B}^d(R)} 
        2\|b\left(x_t, t\right)\|^2 +
        2 d \left(\frac{2L}{c}\right)^2  dP_{x_t} \leq
        \\ \leq  
        \int_{x_t \notin \mathrm{B}^d(R)} 2\|  \nabla \ln  U_t f(x_t)\|^2 +
        8 d \left(\frac{L}{c}\right)^2  dP_{x_t} = \int_{x_t \notin \mathrm{B}^d(R)} 2\left\|  \frac{\nabla U_t f(x_t)}{ U_t f(x_t)}\right\|^2  +
        8 d \left(\frac{L}{c}\right)^2  dP_{x_t} \leq
        \\ \leq \int_{X_t \notin \mathrm{B}^d(R)} 2  \frac{L}{c}^2  +
        8 d \left(\frac{L}{c}\right)^2  dP_{x_t} \leq 9d\frac{L^2}{c^2}    P \left\{ \sup_{t\in [0,1]}\|x_t\| \geq R\right\} \leq \frac{9 d L^2}{c^2} \cdot \frac{\sqrt{d}+L / c}{R}
    \end{aligned}
\end{align*}

Choosing $R$ large enough to guarantee $T_2 \leq \varepsilon$ and putting everything together, we obtain $D(\boldsymbol{\mu} \| \widehat{\boldsymbol{\mu}}) \leq \varepsilon$. Therefore, $D(\mu \| \widehat{\mu}) \leq D(\boldsymbol{\mu} \| \widehat{\boldsymbol{\mu}}) \leq \varepsilon$ by the data processing inequality.
\end{proof}

Finally, we would like to highlight what happens when we sample $\hat{x}_0 \sim \gN(0,1)$ rather than $p_T$. Whilst our results are done for $t \in [0, 1]$ one can see that the overall approximation results will hold for  $t \in [0, T]$.

\begin{repremark}{rem:approx}
    Assuming $\pi$ satisfies a logarithmic Sobolev inequality we extend the time domain to $t\in [0,T]$ and sampling $\hat{x}_0 \sim \gN(0,I)$ approximately, it follows that $D(\mu||\hat{\mu}) \leq  e^{-T}\KL(\pi || \gN(0,1)) + T\epsilon$
\end{repremark}

\begin{proof}

First, we remark that the estimation results and the results in Corollary \ref{col:est} apply to the $t \in [0,T]$ setting, however, they will introduce a polynomial dependency in $T$ for the size of the network.
    
As in the above proof, we apply the Girsanov theorem to control the path KL, however here, the starting distributions of the two Ito processes are no longer the same thus, we get an extra term from the chain rule:
\begin{align}
    \KL(\boldsymbol{\mu} \| \widehat{\boldsymbol{\mu}})&=\KL(p_T || \gN(0,1)) +\frac{1}{2} \int_0^T \mathbf{E}\left\|b\left(x_t, t\right)-\widehat{b}\left(x_t, t\right)\right\|^2 \mathrm{~d} t  \\
    &\leq \KL(p_T || \gN(0,1)) +T \epsilon \\
    &\leq e^{-T}\KL(\pi || \gN(0,1)) +T\epsilon 
\end{align}
Where the final inequality follows from Theorem 5.2.1 in \cite{bakry2014analysis} under the assumption that $\pi$ satisfies a log-Sobolev inequality. This completes the circle and fully extends Theorem 3.1 from \cite{tzen2019theoretical} to our denoising diffusion setting. 

\end{proof}

Finally, note that if we assume that $\operatorname{supp} \pi \subseteq \mathrm{B}^d(R)$ from Theorem 2 of \cite{chen2022sampling} it follows that:
\begin{align}
    \mathrm{TV}\left(\law \hat{x}_t, \pi\right) \leq \gO \left( {\sqrt{\mathrm{KL}\left(\pi \| \gN(0, I)\right)} \exp (-T)}+ {\epsilon \sqrt{T}} \right) .
\end{align}
This result complements Corollary \ref{col:est} very nicely as unlike \cite{chen2022sampling} we no longer require assuming an $\epsilon$ error on the score but instead prove such error can be attained.

\section{PBM Transition Density}\label{apdx:PBM}

As PBM is a linear SDE we know its transition densities are Gaussian thus finding its first and second moments fully determines it.

\subsection{Mean}

Taking expectations on the solution to the PBM-SDE yields an ODE for the mean of the transition density:

$$
\frac{d \mu_t}{dt} = \left(\frac{d \alpha_t}{dt}\right)  \frac{\mu_t }{\alpha_T-\alpha_t} 
$$

separating variables:

$$
\frac{1}{\mu_t}d \mu_t =  \frac{1 }{\alpha_T-\alpha_t} d \alpha_t
$$

integrating both sides:

$$
\ln \frac{\mu_t}{\mu_s} = \ln \frac{\alpha_T-\alpha_t}{\alpha_T-\alpha_s}
$$

thus:
$$
\mu_t = \mu_s \frac{\alpha_T-\alpha_t}{\alpha_T-\alpha_s}
$$
and at $s=0$:
$$
\mu_t = x \frac{\alpha_T-\alpha_t}{\alpha_T-\alpha_0}
$$

\subsection{Variance}

Applying Ito's Lemma to the PBM-SDE $z_t=x_t^2$ yields,
\begin{align}
\mathrm{d}z_t = \left(-\left(\frac{d \alpha_t}{\mathrm{d}t}\right) \frac{2z_t}{\alpha_T-\alpha_t} + \left(\frac{\mathrm{d} \alpha_t}{\mathrm{d}t}\right) \right)dt + 2\left(\frac{d \alpha_t}{\mathrm{d}t}\right)^{1/2}  x_t \mathrm{d}W_t
\end{align}
taking expectations and using the martingale property we have: 
\begin{align}
\frac{d \mu_z(t)}{dt} = \left(\frac{d \alpha_t}{dt}\right)\left(1-\frac{2\mu_z(t)}{\alpha_T-\alpha_t}\right)
\end{align}

As before let us compute the integrating factor : 

$$
e^{-\int_s^t\left(\frac{d \alpha_\tau}{d\tau}\right) \frac{2Z_\tau}{\alpha_T-\alpha_\tau} d\tau} = \left(\frac{\alpha_T-\alpha_s }{\alpha_T-\alpha_t}\right)^2
$$

thus:

\begin{align}
\left(\frac{d \alpha_t}{dt}\right)  \left(\frac{\alpha_T-\alpha_s }{\alpha_T-\alpha_t}\right)^2= \frac{d(((\alpha_T-\alpha_s)^2 /(\alpha_T-\alpha_t)^2))\mu_z(t))}{dt} 
\end{align}

\begin{align}
\int_s^t \left(\frac{d \alpha_\tau}{d\tau}\right)  \left(\frac{\alpha_T-\alpha_s }{\alpha_T-\alpha_\tau}\right)^2 d\tau = \left(\frac{\alpha_T-\alpha_s }{\alpha_T-\alpha_t}\right)^2\mu_z(t) + \mu_z(s)
\end{align}

\begin{align}
\int_s^t  \left(\frac{\alpha_T-\alpha_s }{\alpha_T-\alpha_\tau}\right)^2 d \alpha_\tau = \left(\frac{\alpha_T-\alpha_s }{\alpha_T-\alpha_t}\right)^2\mu_z(t) + \mu_z(s)
\end{align}

\begin{align}
 (\alpha_T-\alpha_s)^2\left(\left(\frac{1}{\alpha_T-\alpha_t}\right)- \left(\frac{1}{\alpha_T-\alpha_s}\right)\right) = \left(\frac{\alpha_T-\alpha_s }{\alpha_T-\alpha_t}\right)^2\mu_z(t) + \mu_z(s)
\end{align}

rearranging:

\begin{align}
\left({\alpha_T-\alpha_t}\right)- \frac{(\alpha_T-\alpha_t)^2}{\alpha_T-\alpha_s} -\left(\frac{\alpha_T-\alpha_t }{\alpha_T-\alpha_s}\right)^2 \mu_z(s) = \mu_z(t) 
\end{align}

Now using $\mathrm{var}(X) = \E[X^2]- \E[X]^2$ give the desired result.

\begin{align}
\mathrm{var}(z_t) = \left({\alpha_T-\alpha_t}\right)- \frac{(\alpha_T-\alpha_t)^2}{\alpha_T-\alpha_s}  =  \frac{(\alpha_T-\alpha_t) (\alpha_t - \alpha_s)}{\alpha_T-\alpha_s} 
\end{align}

\subsection{Transition density}

Using the results from the previous two sections and $s=0$ we have:

$$
p(x_t | x_0) = \mathcal{N}\left(x_t\Bigg| \frac{\alpha_T-\alpha_t }{\alpha_T-\alpha_0} x_0  , \frac{(\alpha_T-\alpha_t) (\alpha_t - \alpha_0)}{\alpha_T-\alpha_0}  \right).
$$

\begin{figure}[t]
\centering
\begin{subfigure}{\textwidth} 
  \centering
  \includegraphics[width=\linewidth]{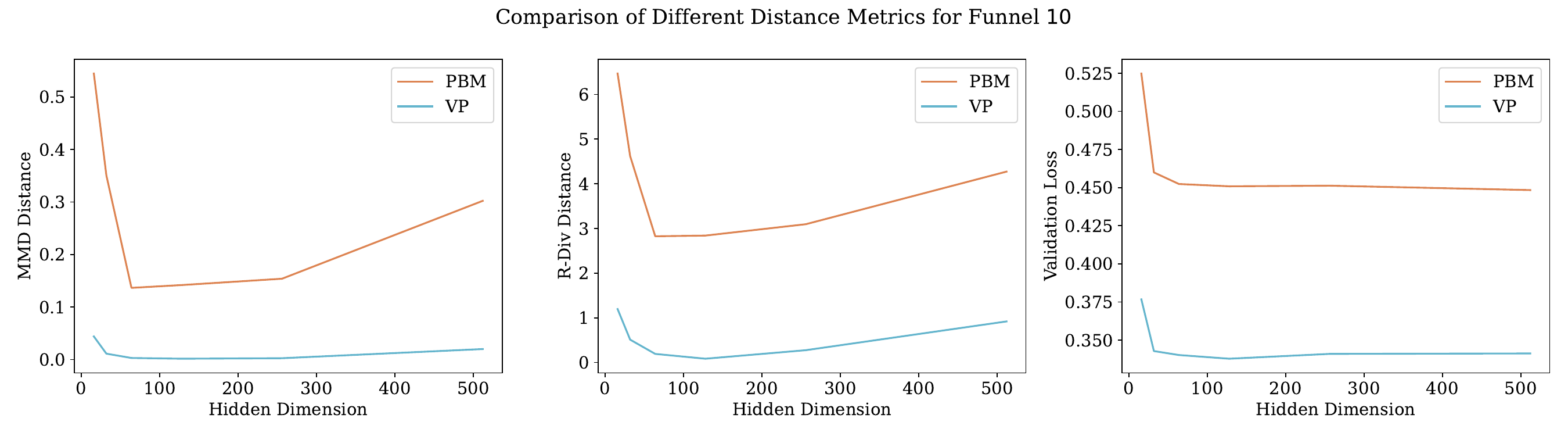} 
  \caption{Distances between $\pi$ and $p^{\mathrm{model}}_\theta$ at time $T$ over a $10$-dimensional Funnel, with results obtained from $3$ different seeds. The $x$-axis represents various hidden layer dimensions}
  \label{fig:funnel_10}
\end{subfigure}

\begin{subfigure}{\textwidth} 
  \centering
  \includegraphics[width=\linewidth]{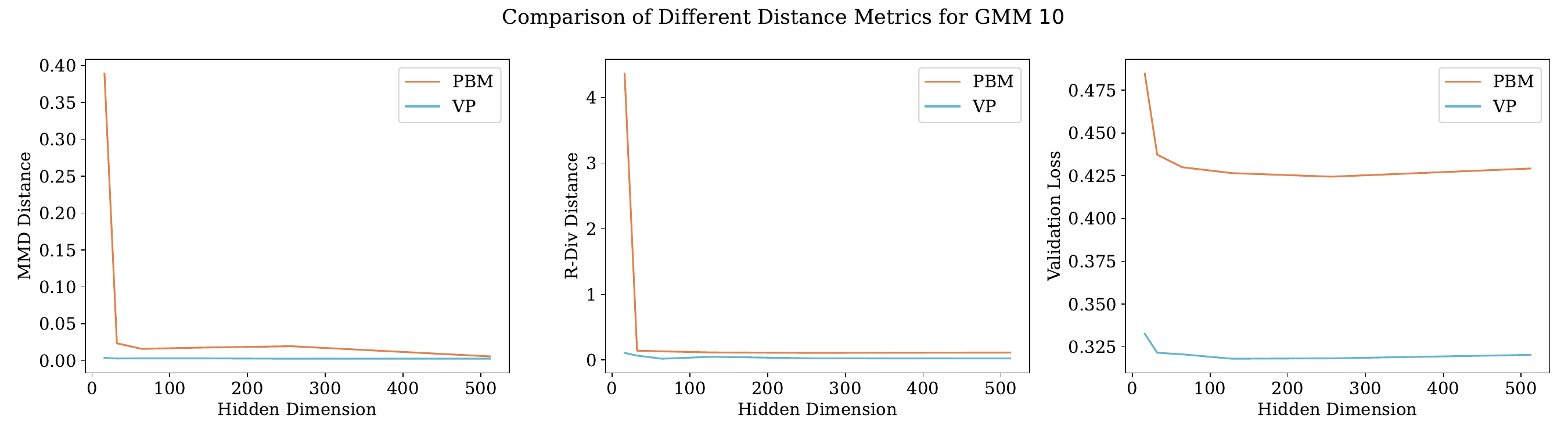} 
  \caption{Distances between $\pi$ and $p^{\mathrm{model}}_\theta$ at time $T$ over a $10$-dimensional GMM-10, with results obtained from $3$ different seeds. The $x$-axis represents various hidden layer dimensions}
  \label{fig:gmm_10}
\end{subfigure}
\label{fig:overall}
\end{figure}

\section{Sinthetic Experimental Details} \label{appdx:sim}

We employed a neural network architecture consisting of 5 MLP layers ReLU activation functions and dropout set to 0. The learning rate was set to 0.00001, and we conducted training over 100 epochs for each model in our study. We utilized the Adam optimizer along with a LambdaLR scheduler.

The datasets were divided into training, validation, and testing sets. The training set consisted of 100,000 samples, the validation set consisted of 20,000 samples, and the testing set consisted of 10,000 samples. To evaluate model performance, we computed Maximum Mean Discrepancy (MMD) and r-divergence between samples generated by the trained model at the final time step and samples from the testing set.

\subsection{Noise Schedule}

For both VP-SDE and PBM-SDE we use the following linear noise schedule:
\begin{align}
    \beta_t = \frac{\mathrm{d} \alpha_t}{\mathrm{d}t} = \beta_{\mathrm{min}} \frac{(T- t)}{T}  + \beta_{\mathrm{max}} \frac{t}{T}
\end{align}
with $T=1$ , $\beta_{\mathrm{min}}= 0.1$, $\beta_{\mathrm{max}} = 20$.
\subsection{Score estimation across network size}
\label{appx:dim_10}
For this experiment, we fixed $d=10$, and varied the network width across $4, 16, 32, 64, 128, 256, 512$ for both GMM-10 and Funnel. The results obtained are in line with those in Section \ref{sec:increase_hid_dim}.

\subsection{MMD and R-divergence Details}

We use MMD-Fuse \citep{biggs2023mmd}\footnote{\url{https://github.com/antoninschrab/mmdfuse}} codebase to compute MMD and a Laplace kernel.

For the R-divergence, we use a standard Gaussian kernel and a Scott bandwidth estimator \citep{scott1979optimal} using the Scipy library \citep{2020SciPy}.


\section{Image Datasets Experimental Details}\label{app:image_data_details}

The used network was a convolutional neural network (CNN) tailored for image processing tasks, specifically optimized for images sized $28\times28$ pixels with three color channels (RGB). It consisted of multiple layers of convolutional and residual blocks, featuring a total of $32$ channels within the convolutional layers and incorporating one residual block. The architecture integrated the principle of channel multiplication, sequentially scaling the number of channels in each layer by factors of $1, 2$, and $2$. The model is also utilizing residual blocks for both upscaling and downscaling operations. The total number of parameters is $1.34$ million.

\subsection{Noise Schedule}
For both VP-SDE and PBM-SDE we use the following linear noise schedule:
\begin{align}
    \beta_t = \frac{\mathrm{d} \alpha_t}{\mathrm{d}t} = \beta_{\mathrm{min}} \frac{(T- t)}{T}  + \beta_{\mathrm{max}} \frac{t}{T},
\end{align}
with values:
\begin{itemize}
    \item for VP-SDE: $T=1$ , $\beta_{\mathrm{min}}= 0.0001$, $\beta_{\mathrm{max}} = 20$.
    \item for PBM: $T=2$ , $\beta_{\mathrm{min}}= 0.001$, $\beta_{\mathrm{max}} = 1$.
\end{itemize}
\subsection{FID Details}
We use \cite{jiralerspong2023feature}\footnote{\url{https://github.com/marcojira/fld}} codebase to compute the FID metric over the test dataset, employing a sample size of $1000$ and the CLIP feature extractor.

\subsection{Samples}

\begin{figure}[h]
    \centering
    \includegraphics[width=0.6\linewidth]{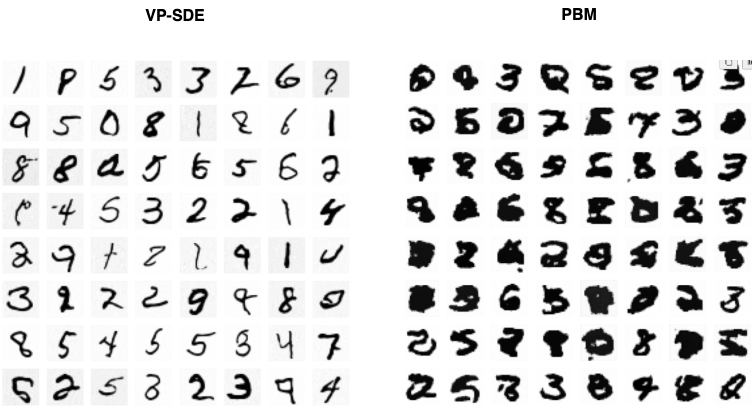}
    \caption{MNIST samples}
    \label{fig:minst_samples}
\end{figure}

\begin{figure}[h]
    \centering
    \includegraphics[width=0.6\linewidth]{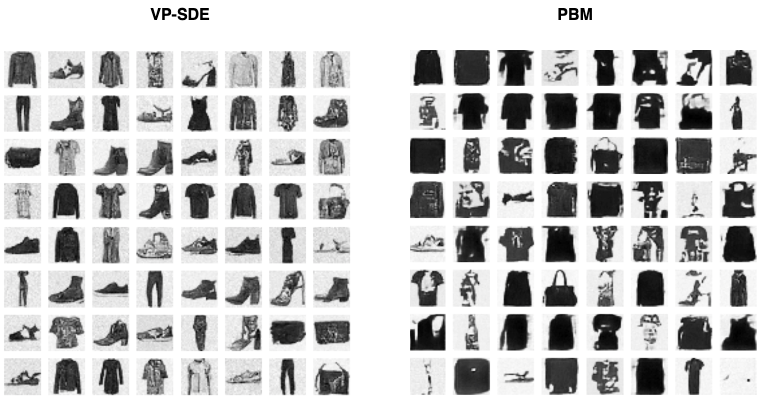}
    \caption{Fashion-MNIST samples}
    \label{fig:fashion_minst_samples}
\end{figure}

\begin{figure}[h]
    \centering
    \includegraphics[width=0.6\linewidth]{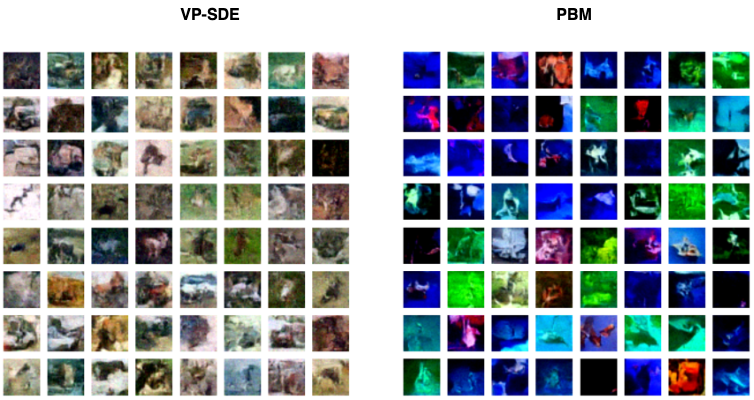}
    \caption{CIFAR-10 samples}
    \label{fig:cifar10_samples}
\end{figure}

\end{document}